\documentclass[journal]{IEEEtran}

\pdfoutput=1

\usepackage{cite}
\usepackage[pdftex]{graphicx}
\usepackage{amsmath}
\usepackage{amsfonts}
\usepackage{bm}
\usepackage{algorithm}
\usepackage{algpseudocode}
\usepackage{array}
\usepackage[caption=false, font=footnotesize]{subfig}
\usepackage{dblfloatfix}
\usepackage{url}

\graphicspath{{./figures/}{./figures/set1/}{./figures/set2/}{./figures/set3/}{./figures/set4/}{./figures/set5/}{./figures/set6/}{./figures/set7/}}

\algnewcommand\Input{ \item[\textbf{Input:}] }
\algnewcommand\Output{ \item[\textbf{Output:}] }

\DeclareMathOperator*{\argmin}{arg \! \min}
\DeclareMathOperator*{\argmax}{arg \! \max}
\DeclareMathOperator*{\diag}{diag}
\newcommand{\prox}[1]{\text{prox} \tiny #1}

\begin{document}

\title{Light Field Super-Resolution\\ Via Graph-Based Regularization}

\author{Mattia~Rossi~and~Pascal~Frossard \\
	\'{E}cole Polytechnique F\'{e}d\'{e}rale de Lausanne \\
	mattia.rossi@epfl.ch, pascal.frossard@epfl.ch}

\maketitle

\begin{abstract}
Light field cameras capture the 3D information in a scene with a single exposure.
This special feature makes light field cameras very appealing for a variety of applications: from post-capture refocus,
to depth estimation and image-based rendering.
However, light field cameras suffer by design from strong limitations in their spatial resolution, which should therefore be augmented
by computational methods.
On the one hand, off-the-shelf single-frame and multi-frame super-resolution algorithms are not ideal for light field data,
as they do not consider its particular structure.
On the other hand, the few super-resolution algorithms explicitly tailored for light field data exhibit significant limitations,
such as the need to estimate an explicit disparity map at each view.
In this work we propose a new light field super-resolution algorithm meant to address these limitations.
We adopt a multi-frame alike super-resolution approach, where the complementary information in the different light field views
is used to augment the spatial resolution of the whole light field.
We show that coupling the multi-frame approach with a graph regularizer, that enforces the light field structure via nonlocal self similarities,
permits to avoid the costly and challenging disparity estimation step for all the views.
Extensive experiments show that the new algorithm compares favorably to the other state-of-the-art methods for light field
super-resolution, both in terms of PSNR and visual quality.
\end{abstract}

\IEEEpeerreviewmaketitle

\section{Introduction}

We live in a 3D world but the pictures taken with traditional cameras can capture just 2D projections of this reality.
The \textit{light field} is a model that has been originally introduced in the context of image-based rendering with the purpose
of capturing richer information in a 3D scene \cite{levoy_light_1996} \cite{gortler_lumigraph_1996}.
The light emitted by the scene is modeled in terms of rays, each one characterized by a direction and a radiance value.
The light field function provides, at each point in space, the radiance from a given direction.
The rich information captured by the light field function could be used in many applications, from post-capture refocus to
depth estimation or virtual reality.

However, the light field is a theoretical model: in practice the light field function has to be properly sampled, which is a challenging task.
A straightforward but hardware-intensive approach relies on camera arrays \cite{wilburn_high_2005}.
In this setup, each camera records an image of the same scene from a particular position and the light field takes the form of an array of views.
More recently, the development of the first commercial light field cameras \cite{lytro_inc} \cite{raytrix_gmbh} has made light field sampling
more accessible.
In light field cameras, a micro lens array placed between the main lens and the sensor permits to virtually partition
the main lens into sub-apertures, whose images are recorded altogether in a single exposure \cite{ng_light_2005} \cite{perwass_single_2012}.
As a consequence, a light field camera behaves as a compact camera array, providing multiple simultaneous images of a 3D scene
from slightly different points of view.

Even if light field cameras become very appealing, they still face the so called \textit{spatio-angular resolution tradeoff}.
Since the whole array of views is captured by a single sensor, a dense sampling of the light field in the angular domain (i.e., a large
number of views) necessarily translates into a sparse sampling in the spatial domain (i.e., low resolution views) and vice versa.
A dense angular sampling is at the basis of any light field application, as the 3D information provided by the light field data comes
from the availability of different views.
It follows that the angular sampling cannot be excessively penalized to favor spatial resolution.
Moreover, even in the limit scenario of a light field with just two views, the spatial resolution of each one may be reduced to half of
the sensor \cite{ng_light_2005}, which still happens to be a dramatic drop in the resolution.
Consequently, the light field views exhibit a significantly lower resolution than images from traditional cameras, and many light field
applications, such as depth estimation, happen to be very challenging on low spatial resolution data.
The design of spatial super-resolution techniques, aiming at increasing the view resolution, is therefore crucial in order to fully
exploit the potential of light field cameras.

In this work, we propose a new light field super-resolution algorithm that provides a global solution that augments the resolution
of all the views together, without an explicit a priori disparity estimation step.
In particular, we propose to cast \textit{light field spatial super-resolution} into a global optimization problem, whose objective function
is designed to capture the relations between the light field views.
The objective function comprises three terms.
The first one enforces data fidelity, by constraining each high resolution view to be consistent with its low resolution counterpart.
The second one is a warping term, which gathers for each view the complementary information encoded in the other ones.
The third one is a graph-based prior, which regularizes the high resolution views by enforcing smoothness along the light field epipolar
lines that define the light field structure.
These terms altogether form a quadratic objective function that we solve iteratively with the proximal point algorithm.
The results show that our algorithm compares favorably to state-of-the-art light field super-resolution algorithms, both visually and
in terms of reconstruction error.

The article is organized as follows.
Section \ref{sec:related_work} presents an overview of the related literature.
Section \ref{sec:light_field_structure} formalizes the light field structure.
Section \ref{sec:problem_form} presents our problem formulation and carefully analyzes each of its terms.
Section \ref{sec:super_res_algo} provides a detailed description of our super-resolution algorithm,
and Section \ref{sec:complexity} analyses its computational complexity.
Section \ref{sec:experiments} is dedicated to our experiments.
Finally, Section \ref{sec:conclusions} concludes the article.

\section{Related work \label{sec:related_work}}

The super-resolution literature is quite vast, but it can be divided mainly into two areas: single-frame and multi-frame super-resolution methods.
In single-frame super-resolution, only one image from a scene is provided, and its resolution has to be increased.
This goal is typically achieved by learning a mapping from the low resolution data to the high resolution one, either on an external training set
\cite{jianchao_yang_2010} \cite{xinbo_joint_2012} \cite{dong_learning_2014} or on the image itself \cite{glasner_super_2009} \cite{bevilacqua_single_2014}.
Single-frame algorithms can be applied to each light field view separately in order to augment the resolution of the whole light field,
but this approach would neither exploit the high correlation among the views, nor enforce the consistency among them.

In the multi-frame scenario, multiple images of the same scene are used to increase the resolution of a target image.
To this purpose, all the available images are typically modeled as translated and rotated versions of the target one \cite{irani_1991} \cite{farsiu_fast_2004}.
The multi-frame super-resolution scenario resembles the light field one, but its global image warping model does not
fit the light field structure.
In particular, the different moving speeds of the objects in the scene across the light field views, which encode their different depths,
cannot be captured by a global warping model.
Multi-frame algorithms employing more complex warping models exist, for example in video super-resolution \cite{mitzel_video_2009}
\cite{unger_convex_2010}, yet the warping models do not exactly fit the geometry of light field data and their construction is computationally demanding.
In particular, multi-frame video super-resolution involves two main steps, namely optical flow estimation, which finds correspondences
between temporally successive frames, and eventually a super-resolution step that is built on the optical flow.

In the light field representation, the views lie on a two-dimensional grid with adjacent views sharing a constant baseline under the assumption
of both vertical and horizontal registration.
As a consequence, not only the optical flow computation reduces to disparity estimation, but the disparity map at one view
determines its warping to every other view in the light field, in the absence of occlusions.
In \cite{wanner_spatial_2012} Wanner and Goldluecke build over these observations to extract the disparity map at each
view directly from the epipolar line slopes with the help of a structure tensor operator.
Then, similarly to multi-frame super-resolution, they project all the views to the target one within a global optimization formulation endowed
with a \textit{Total Variation (TV)} prior.
Although the structure tensor operator permits to carry out disparity estimation in the continuous domain, this task remains very challenging
at low spatial resolution.
As a result, disparity errors unfortunately translate into significant artifacts in the textured areas and along object edges.
Finally, each view of the light field has to be processed separately to super-resolve the complete light field, which does not permit to fully
exploit the inter-view dependencies.

In another work, Heber and Pock \cite{heber_shape_2014} consider the matrix obtained by warping all the views to a reference one,
and propose to model it as the sum of a low rank matrix and a noise one, where the later describes the noise and occlusions.
This model, that resembles \textit{Robust PCA} \cite{candes_robust_2011}, is primarily meant for disparity estimation at the reference view.
However, the authors show that a slight modification of the objective function can provide the corresponding high resolution view,
in addition to the low resolution disparity map at the reference view.
The algorithm could ideally be applied separately to each view in order to super-resolve the whole light field, but that may not be the ideal
solution to that global problem, due to the high redundancy in estimating all the low resolution disparity maps independently.

In a different framework, Mitra and Veeraraghavan propose a light field super-resolution algorithm based on a learning procedure \cite{mitra_light_2012}.
Each view in the low resolution light field is divided into patches that are possibly overlapping.
All the patches at the same spatial coordinates in the different views form a light field with very small spatial resolution, i.e., a \textit{light field patch}.
The authors assign a constant disparity to each light field patch, i.e., all the objects within the light field patch are assumed to lie
at the same depth in the scene.
A different \textit{Gaussian Mixture Model (GMM)} prior for high resolution light field patches is learnt offline for each discrete disparity value,
and it is then employed within a \textit{MAP} estimator to super-resolve each light field patch with the corresponding disparity.
Due to the reduced dimensionality of each light field patch, and to the closed form solution of the estimator, this approach requires less
online computation than other light field super-resolution algorithms.
However, the offline learning strategy has also some drawbacks:
the dependency of the reconstruction on the chosen training set,
the need for a new training for each super-resolution factor,
and finally the need for a proper discretization of the disparity range, which introduces a tradeoff between the reconstruction quality
and the time required by both the training and the reconstruction steps.
Moreover, the simple assumption of constant disparity within each light field patch leads to severe artifacts at depth discontinuities
in the super-resolved light field views.

The light field super-resolution problem has been addressed within the framework of \textit{Convolutional Neural Networks (CNNs)} too.
In particular, Yoon et al. \cite{yoon_light-field_2017} consider the cascade of two CNNs, the first meant to super-resolve the given
light field views, and the second to synthesize new high resolution views based on the previously super-resolved ones.
However, the first CNN (whose design is borrowed from \cite{dong_learning_2014}) is meant for single-frame super-resolution,
therefore the views are super-resolved independently, without considering the light field structure.

Finally, we note that some authors, e.g., Bishop et al. \cite{bishop_light_2012}, consider the recovery of an all in focus image with full
sensor resolution from the light field camera output.
They refer to this task as light field super-resolution although it is different from the problem considered in this work.
In this article, no light field applications is considered a priori: the light field views are all super-resolved, thus enabling any light field
application to be performed later at a resolution higher than the original one.
Differently from the other light field super-resolution algorithms, ours does not require an explicit a priori disparity estimation step,
and does not rely on a learning procedure.
Moreover, our algorithm reconstructs all the views jointly, provides homogeneous quality across the reconstructed views,
and preserves the light field structure.

\begin{figure}
	\centering
	\includegraphics[width=0.99 \linewidth]{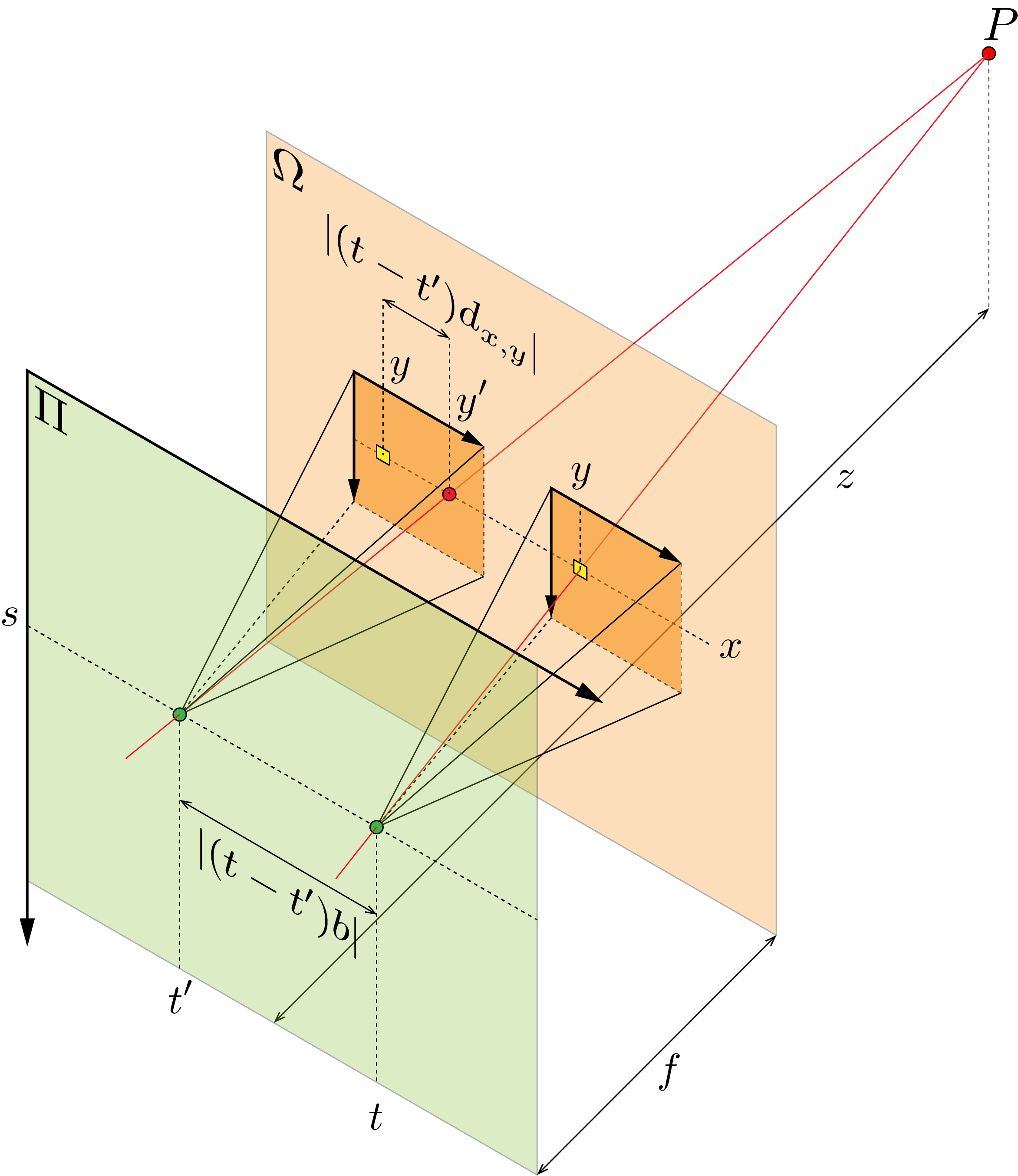}
	\caption{Light field sampling in the two plane parametrization.
	The light field is ideally sampled through an ${M \times M}$ array of pinhole cameras.
	The pinhole cameras at coordinates ${(s,t)}$ and ${(s,t')}$ in the camera array are represented as two pyramids,
	with their apertures denoted by the two green dots on the plane $\Pi$, and their ${N \times N}$ sensors
	represented by the two orange squares on the plane $\Omega$.
	The distance between the two planes is the \textit{focal length} $f$.
	The distance between the apertures of horizontally or vertically adjacent views in the $M \times M$ array is the
	\textit{baseline} $b$, hence the distance between the two green dots on plane $\Pi$ is ${|(t - t') b|}$.
	The small yellow squares in the two sensors denote pixel ${(x,y)}$.
	Pixel ${(x,y)}$ of camera ${(s,t)}$ captures one of the light rays (in red) emitted by a point $P$ at depth $z$ in the scene.
	The disparity associated to pixel ${(x,y)}$ of camera ${(s,t)}$ is ${d _{x,y}}$, therefore the projection of $P$
	on the sensor of camera ${(s,t')}$ lies at ${(x, y') = (x, y + \left( t - t' \right) d _{x,y})}$.
	The intersection coordinate ${(x, y')}$ is denoted by a red spot, as it does not necessarily correspond to a pixel,
	since ${d _{x,y}}$ is not necessarily integer.}
	\label{fig:two_planes}
\end{figure}
%

\section{Light field structure} \label{sec:light_field_structure}

In the light field literature, it is common to parametrize the light rays from a 3D scene by the coordinates of their intersection with
two parallel planes, typically referred to as the \textit{spatial plane} $\Omega$ and the \textit{angular plane} $\Pi$.
Each light ray is associated to a radiance value, and a pinhole camera with its aperture on the plane $\Pi$ and its sensor on the plane
$\Omega$ can record the radiance of all those rays accommodated by its aperture.
This is represented in Figure~\ref{fig:two_planes}, where each pinhole camera is represented as a pyramid, with its vertex and basis
representing the camera aperture and sensor, respectively.
In general, an array of pinhole cameras can perform a regular sampling of the angular plane $\Pi$, therefore the sampled light field takes
the form of a set of images captured from different points of view.
This is the sampling scheme approximated by both camera arrays and light field cameras.

In the following we consider the light field as the output of an ${M \times M}$ array of pinhole cameras, each one equipped with
an ${N \times N}$ pixel sensor.
Each camera (aperture) is identified through the angular coordinate ${(s,t)}$ with ${s,t \in \{ 1, 2, ... , M \}}$, while a pixel within the camera sensor
is identified through the spatial coordinate ${(x,y)}$ with ${x,y \in \{ 1, 2, ... , N \}}$.
The distance between the apertures of horizontally or vertically adjacent cameras is $b$, referred to as the \textit{baseline}.
The distance between the planes $\Pi$ and $\Omega$ is $f$, referred to as the camera \textit{focal length}.
Figure~\ref{fig:two_planes} sketches two cameras of the ${M \times M}$ array.
Within this setup, we can represent the light field as an ${N \times N \times M \times M}$ real matrix $\bm{U}$, with ${\bm{U} (x, y, s, t)}$
the intensity of a pixel with coordinates ${(x, y)}$ in the view of camera ${(s, t)}$.
In particular, we denote the view at ${(s, t)}$ as $\bm{U} _{s, t} \equiv \bm{U} (\cdot, \cdot, s, t) \in \mathbb{R} ^{N \times N}$.
Finally, without lack of generality, we assume that each pair of horizontally or vertically adjacent views in the light field are properly registered.

\begin{figure}
	\centering
	\subfloat[]{%
		\includegraphics[width= 0.95 \linewidth]{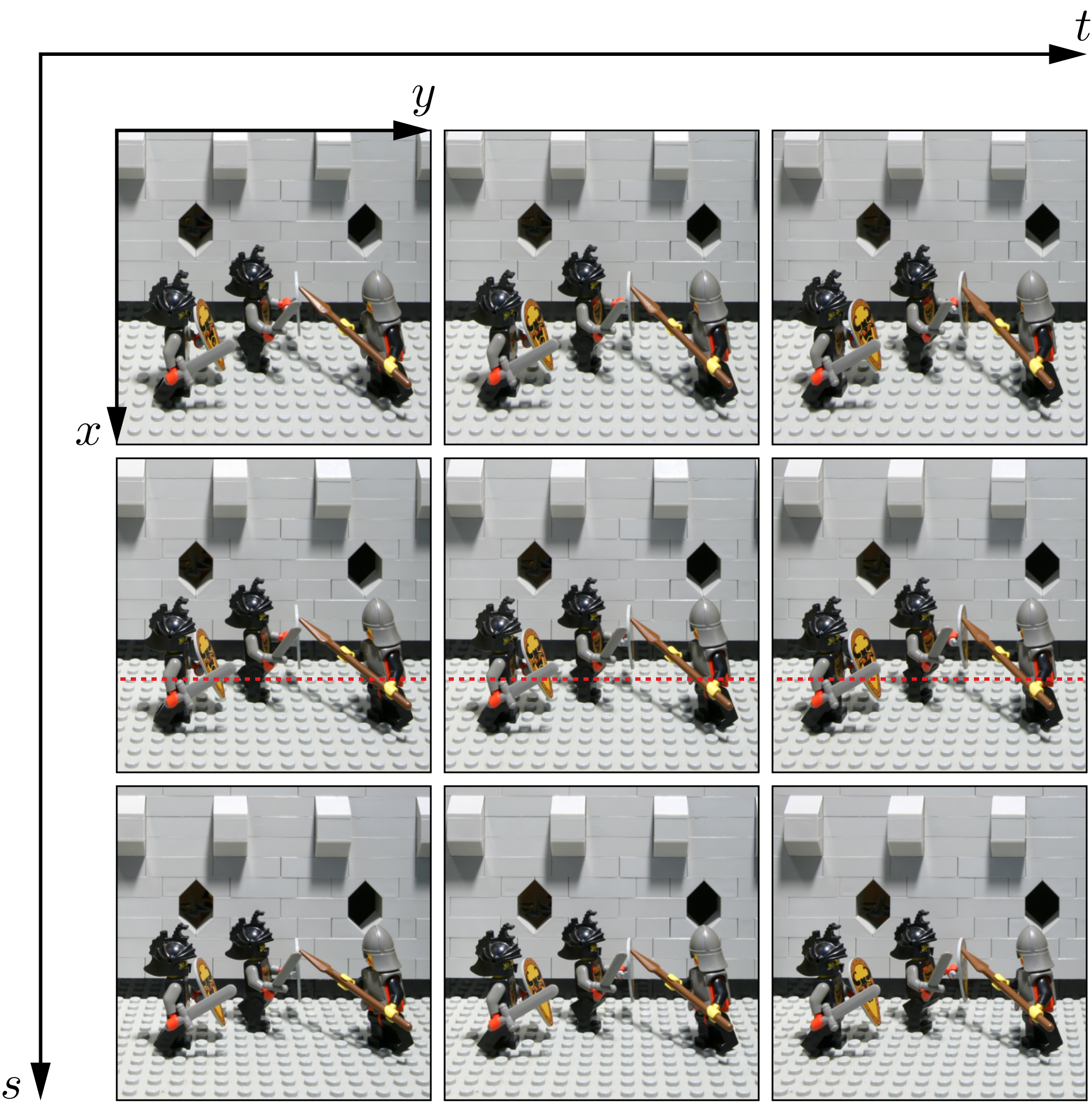}
		\label{fig:light_field_example}
	}
	\vspace{0.05cm}
	\subfloat[]{%
		\includegraphics[width=0.95 \linewidth]{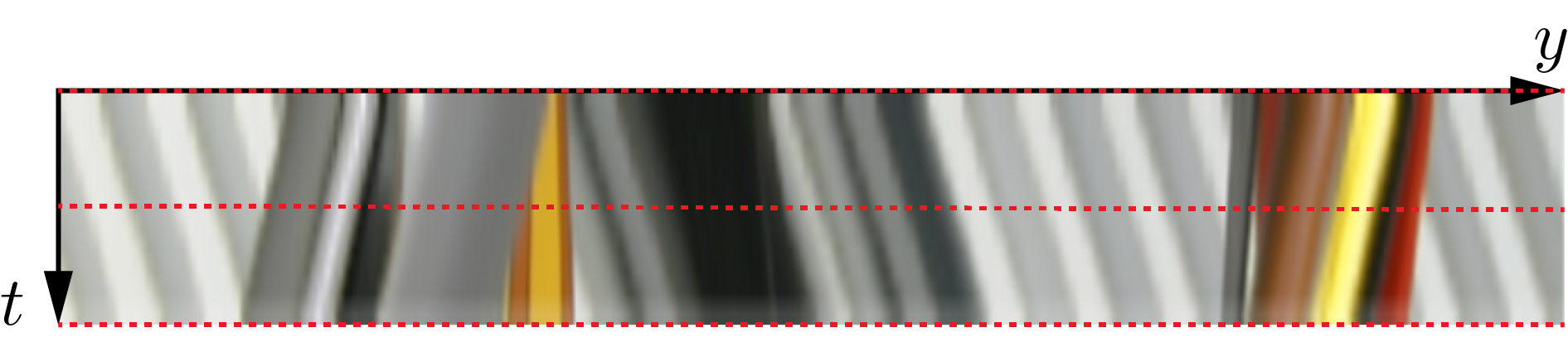}
		\label{fig:epi_example}
	}
	\caption{Example of light field and epipolar image (EPI). Figure (a) shows an array of $3 \times 3$ views, extracted from the
	\texttt{knights} light field (Stanford dataset) which actually consists of an array of $17 \times 17$ views.
	Figure (b) shows an epipolar image from the original $17 \times 17$ \texttt{knights} light field.
	In particular, the \mbox{$t$-th} row in the epipolar image correspond to the row $\bm{U} _{9, t} (730, \cdot)$.
	The top, central, and bottom red dashed rows in (b) corresponds to the left, central, and right dashed rows in red in the sample
	views in (a), respectively.}
\end{figure}

With reference to Figure~\ref{fig:two_planes}, we now describe in more details the particular structure of the light field data.
We consider a point $P$ at depth ${z}$ from $\Pi$, whose projection on one of the cameras is represented by the pixel ${\bm{U} _{s, t} (x, y)}$,
in the right view of Figure~\ref{fig:two_planes}.
We now look at the projection of $P$ on the other views ${\bm{U} _{s, t'}}$ in the same row of the camera array, such as the left view in Figure~\ref{fig:two_planes}.
We observe that, in the absence of occlusions and under the Lambertian assumption\footnote{All the rays emitted by point $P$
exhibit the same radiance.}, the projection of $P$ obeys the following stereo equation:
\begin{align}
	\bm{U} _{s, t} \left( x, y \right) \>
	&= \> \bm{U} _{s, t'}  \left( x, \> y + \left( t - t' \right) d _{x,y} \right) \nonumber \\
	&= \> \bm{U}  _{s, t'}  \left( x, y' \right) \label{eq:onedim_stereo}
\end{align}
where ${d _{x, y} \equiv \bm{D} _{s,t} (x, y) \equiv fb/z}$, with ${\bm{D} _{s, t} \in \mathbb{R} ^{N \times N}}$ the disparity map
of view ${\bm{U} _{s, t}}$ with respect to its left view ${\bm{U} _{s, t - 1}}$.
A visual interpretation of Eq.~(\ref{eq:onedim_stereo}) is provided by the \textit{Epipolar Plane Image (EPI)} \cite{bolles_1987}
in Figure~\ref{fig:epi_example}, which represents a slice ${\bm{E} _{s, x} \equiv \bm{U} (x, \cdot, s, \cdot) ^{\top} \in \mathbb{R} ^{M \times N}}$ of the light field.
This is obtained by stacking the $x$-th row from each view ${\bm{U} _{s, t'}}$, with ${t' = 1, 2, \ldots, M}$, on top of each other.
This procedure leads to a clear line pattern, as the projection ${\bm{U} _{s, t} (x, y) = \bm{E} _{s, x} (t, y)}$ of point $P$ on the view at ${\bm{U} _{s, t}}$
is the pivot of a line hosting all its projections on the other views ${\bm{U} _{s, t'}}$.
In particular, the line slope depends on ${d _{x, y}}$, hence on the depth of the point $P$ in the scene.
We stress out that, although ${\bm{U} _{s, t} (x, y)}$ is a pixel in the captured light field, all its projections ${\bm{U} _{s, t'} (x, y')}$
do not necessarily correspond to actual pixels in the light field views, as ${y'}$ may not be integer.
We finally observe that Eq.~(\ref{eq:onedim_stereo}) can be extended to the whole light field:
\begin{align}
	\bm{U} _{s, t} (x, y) \>
	&= \> \bm{U} _{s', t'} (x + \left( s - s' \right) d _{x,y}, \> y + \left( t - t' \right) d _{x,y}) \nonumber \\
	&= \> \bm{U} _{s', t'} \left( x', y' \right). \label{eq:twodim_stereo}
\end{align}
We refer to the model in Eq.~(\ref{eq:twodim_stereo}) as the \textit{light field structure}.

Later on, for the sake of clarity, we will denote a light field view either by its angular coordinate ${(s, t)}$ or by its linear coordinate
${k = ((t - 1) M + s) \in \{ 1, 2, \ldots, M ^{2} \}}$.
In particular, we have ${\bm{U} _{s, t} = \bm{U} _{k}}$ where ${\bm{U} _{k}}$ is the \mbox{$k$-th} view encountered when visiting
the camera array in column major order.
Finally, we also handle the light field in a vectorized form, with the following notation:
\begin{itemize}
\item ${\bm{u} _{s, t} = \bm{u} _{k} \in \mathbb{R} ^{N ^{2}}}$ is the vectorized form of view ${\bm{U} _{s, t}}$,
\item ${\bm{u} = [ \bm{u} _{1} ^{\top}, \bm{u} _{2} ^{\top}, \ldots, \bm{u} _{M ^{2}} ^{\top} ] ^{\top} \in \mathbb{R} ^{N ^{2} M ^{2}}}$,
\end{itemize}
where the vectorized form of a matrix is simply obtained by visiting its entries in column major order.

%

\section{Problem formulation} \label{sec:problem_form}

The light field (spatial) super-resolution problem concerns the recovery of the high resolution light field $\bm{U}$ from its low resolution
counterpart $\bm{V}$ at resolution $(N / \alpha) \times (N / \alpha) \times M \times M$, with ${\alpha \in \mathbb{N}}$ the super-resolution factor.
Equivalently, we aim at super-resolving each view ${\bm{V} _{s, t} \in \mathbb{R} ^{(N / \alpha) \times (N / \alpha)}}$ to get its high
resolution version ${\bm{U} _{s, t} \in \mathbb{R} ^{N \times N}}$.
In order to recover the high resolution light field from the low resolution data, we propose to minimize the following objective function:
\begin{equation} \label{eq:objective_function}
\begin{gathered}
	\bm{u} ^{*} \> \in \> \argmin _{\bm{u}} \> \vphantom{\sum _{k}} \mathcal{F} \left( \bm{u} \right) \\
		\text{with} \quad
		\mathcal{F} \left( \bm{u} \right) \> \equiv \>
		\mathcal{F} _{1} \left( \bm{u} \right) \> + \>
		\lambda _{2} \mathcal{F} _{2} \left( \bm{u} \right) \> + \>
		\lambda _{3} \mathcal{F} _{3} \left( \bm{u} \right)
\end{gathered}
\end{equation}
where each term describes one of the constraints about the light field structure and the multipliers $\lambda _{2}$ and $\lambda _{3}$
balance the different terms.
We now analyze each one of them separately.

Each pair of high and low resolution views have to be consistent, and we model their relationship as follows:
\begin{equation} \label{eq:hr2lr}
	\bm{v} _{k} \> = \> \bm{S} \bm{B} \> \bm{u} _{k} \> + \> \bm{n} _{k}
\end{equation}
where $\bm{B} \in \mathbb{R} ^{N ^{2} \times N ^{2}}$ and ${\bm{S} \in \mathbb{R} ^{(N / \alpha) ^{2} \times N ^{2}}}$ denote a blurring
and a sampling matrix, respectively, and the vector ${\bm{n} _{k} \in \mathbb{R} ^{(N / \alpha) ^{2}}}$ captures possible inaccuracies
of the assumed model.
The first term in Eq.~(\ref{eq:objective_function}) enforces the constraint in Eq.~(\ref{eq:hr2lr}) for each high resolution and low resolution
view pair, and it is typically referred to as the \textit{data fidelity term}:
\begin{equation} \label{eq:first_term}
	\mathcal{F} _{1} \left( \bm{u} \right) \> \equiv \> \sum _{k} \| \bm{S} \bm{B} \> \bm{u} _{k} \> - \> \bm{v} _{k} \| _{2} ^{2}.
\end{equation}

Then, the various low resolution views in the light field capture the scene from slightly different perspectives, therefore details dropped
by digital sensor sampling at one view may survive in another one.
Gathering at one view all the complementary information from the others can augment its resolution.
This can be achieved by enforcing that the high resolution view ${\bm{u} _{k}}$ can generate all the other low resolution views
${\bm{v} _{k'}}$ in the light field, with ${k' \neq k}$.
For every view ${\bm{u} _{k}}$ we thus have the following model:
\begin{equation} \label{eq:hr2lr_warp}
	\bm{v} _{k'} \> = \> \bm{S} \bm{B} \bm{F} _{k} ^{k'} \> \bm{u} _{k} + \bm{n} _{k} ^{k'}, \quad \forall \> k' \neq k
\end{equation}
where the matrix ${\bm{F} _{k} ^{k'} \in \mathbb{R} ^{N ^{2} \times N ^{2}}}$ is such that ${\bm{F} _{k} ^{k'} \bm{u} _{k} \simeq \bm{u} _{k'}}$
and it is typically referred to as a \textit{warping matrix}.
The vector ${\bm{n} _{k} ^{k'}}$ captures possible inaccuracies of the model, such as the presence of pixels of ${\bm{v} _{k'}}$ that cannot
be generated because they correspond to occluded areas in ${\bm{u} _{k}}$.
The second term in Eq.~(\ref{eq:objective_function}) enforces the constraint in Eq.~(\ref{eq:hr2lr_warp}) for every high resolution view:
\begin{equation} \label{eq:second_term}
	\mathcal{F} _{2} \left( \bm{u} \right) \> \equiv \> \sum _{k} \sum _{k' \in \> \mathcal{N} _{k} ^{+}} \| \bm{H} _{k} ^{k'}
	\left( \bm{S} \bm{B} \bm{F} _{k} ^{k'} \> \bm{u} _{k} \> - \> \bm{v} _{k'} \right) \| _{2} ^{2}
\end{equation}
where the matrix ${\bm{H} _{k} ^{k'} \in \mathbb{R} ^{(N / \alpha) ^{2} \times (N / \alpha) ^{2}}}$ is diagonal and binary, and masks
those pixels of ${\bm{v} _{k'}}$ that cannot be generated due to occlusions in ${\bm{u} _{k}}$, while ${\mathcal{N} _{k} ^{+}}$ denotes
a subset of the views (potentially all) with ${k \notin \mathcal{N} _{k} ^{+}}$.

Finally, a regularizer ${\mathcal{F} _{3}}$ happens to be necessary in the overall objective function of Eq.~(\ref{eq:objective_function}),
as the original problem in Eq.~(\ref{eq:hr2lr}), and encoded in term ${\mathcal{F} _{1}}$, is ill-posed due to the fat matrix $\bm{S}$.
The second term ${\mathcal{F} _{2}}$ can help, but the warping matrices ${\bm{F} _{k} ^{k'}}$ in Eq.~(\ref{eq:second_term}) are not
known exactly, such that the third term ${\mathcal{F} _{3}}$ is necessary.

We borrow the regularizer from \textit{Graph Signal Processing (GSP)} \cite{elmoataz_nonlocal_2008} \cite{shuman_emerging_2013},
and define ${\mathcal{F} _{3}}$ as follows:
\begin{equation} \label{eq:third_term}
	\mathcal{F} _{3} \left( \bm{u} \right) \> \equiv \> \vphantom{\sum} \bm{u} ^{\top} \bm{L} \> \bm{u}
\end{equation}
where the positive semi-definite matrix ${\bm{L} \in \mathbb{R} ^{M ^{2} N ^{2} \times M ^{2} N ^{2}}}$ is the \textit{un-normalized Laplacian}
of a graph designed to capture the light field structure.
In particular, each pixel in the high resolution light field is modeled as a vertex in a graph, where the edges connect each pixel to its
projections on the other views.
The quadratic form in Eq.~(\ref{eq:third_term}) enforces connected pixels to share similar intensity values, thus promoting the light
field structure described in Eq.~(\ref{eq:twodim_stereo}).

In particular, we consider an \textit{undirected} weighted graph ${\mathcal{G} = (\mathcal{V}, \mathcal{E}, \mathcal{W})}$,
with $\mathcal{V}$ the set of graph vertices, $\mathcal{E}$ the edge set, and $\mathcal{W}$ a function mapping each edge
into a non negative real value, referred to as the \textit{edge weight}:
\begin{equation*}
	\mathcal{W}: \mathcal{E} \subseteq \left( \mathcal{V} \times \mathcal{V} \right) \rightarrow \mathbb{R},
	\quad \left( i, j \right) \mapsto \mathcal{W} \left( i, j \right).
\end{equation*}
The vertex ${i \in \mathcal{V}}$ corresponds to the entry ${\bm{u} (i)}$ of the high resolution light field, therefore the graph can be
represented through its \textit{adjacency matrix} $\bm{W} \in \mathbb{R} ^{|\mathcal{V}| \times |\mathcal{V}|}$,
with ${|\mathcal{V}|}$ the number of pixels in the light field:
\begin{equation*}
	\bm{W} \left( i, j \right) \> = \>
	\begin{cases}
		\mathcal{W} \left( i, j \right) \quad & \text{if \hspace{0.1cm}} (i, j) \in \mathcal{E} \\
		0 \quad & \text{otherwise}.
	\end{cases}
\end{equation*}
Since the graph is assumed to be undirected, the adjacency matrix is symmetric: ${\bm{W} (i, j) = \bm{W} (j, i)}$.
We can finally rewrite the term $\mathcal{F} _{3}$ in Eq.~(\ref{eq:third_term}) as follows:
\begin{equation} \label{eq:graph_quad_form}
	\mathcal{F} _{3} \left( \bm{u} \right) \>
	= \> \frac{1}{2} \sum _{i} \sum _{j \sim i} \bm{W} \left( i, j \right) \left( \bm{u} \left( i \right) - \bm{u} \left( j \right) \right) ^{2}
\end{equation}
where ${j \sim i}$ denotes the set of vertices $j$ directly connected to the vertex $i$,
and we recall that the scalar ${\bm{u}(i)}$ is the $i$-th entry of the vectorized light field $\bm{u}$.
In Eq.~(\ref{eq:graph_quad_form}) the term ${\mathcal{F} _{3}}$ penalizes significant intensity variations along highly weighted edges.
A weight typically captures the similarity between vertices, therefore the minimization of Eq.~(\ref{eq:graph_quad_form})
leads to an adaptive smoothing \cite{elmoataz_nonlocal_2008}, ideally along the EPI lines of Figure~\ref{fig:epi_example} in our
light field framework.

Differently from the other light field super-resolution methods, the proposed formulation permits to address the recovery of the whole
light field altogether, thanks to the global regularizer ${\mathcal{F} _{3}}$.
The term ${\mathcal{F} _{2}}$ permits to augment the resolution of each view without recurring to external data and learning procedures.
However, differently from video super-resolution or the light field super-resolution approach in \cite{wanner_spatial_2012},
the warping matrices in ${\mathcal{F} _{2}}$ do not rely on a precise estimation of the disparity at each view.
This is possible mainly thanks to the graph regularizer ${\mathcal{F} _{3}}$, that acts on each view as a denoising term based on
nonlocal similarities \cite{kheradmand_general_2014} but at the same time constraints the reconstruction of all the views jointly,
thus enforcing the full light field structure captured by the graph.

\begin{figure}
	\centering
	\includegraphics[width=0.95 \linewidth]{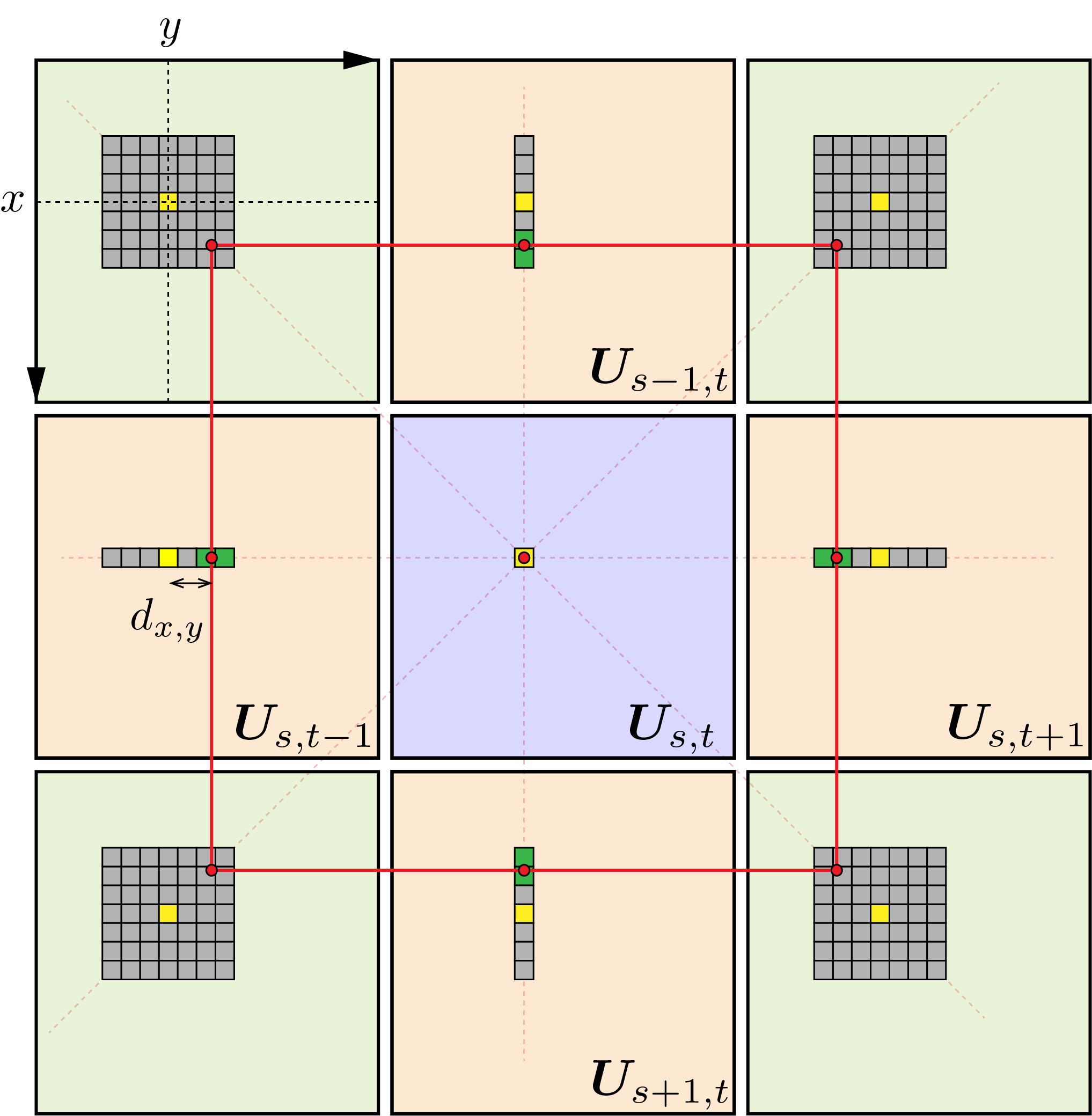}
	\caption{The neighboring views and the square constraint.
	All the squares indicate pixels, in particular, all the yellow pixels lie at the spatial coordinates ${(x, y)}$ in their own view.
	The projection of pixel ${{U} _{s, t} (x, y)}$ on the other eight neighboring views is indicated with a red dot.
	According to Eq.~(\ref{eq:twodim_stereo}) they all lie on a square, highlighted in red.
	The four orange views represent the set ${\mathcal{N} _{k} ^{+}}$, used in the warping matrix construction.
	In each one of these four views, the two pixels employed in the convex combination targeting pixel ${{U} _{s, t} (x, y)}$
	are indicated in green and are adjacent to each other.
	The projection of pixel ${{U} _{s, t} (x, y)}$ lies between the two green pixels, and these two belong to a 1D window indicated in gray.}
	\label{fig:neighborhood}
\end{figure}
%

\section{Super-resolution algorithm \label{sec:super_res_algo}}

We now describe the algorithm that we use to solve the optimization problem in Eq.~(\ref{eq:objective_function}).
We first discuss the construction of the warping matrices of the term ${\mathcal{F} _{2}}$ in Eq.~(\ref{eq:second_term}),
and then the construction of the graph employed in the regularizer ${\mathcal{F} _{3}}$ in Eq.~(\ref{eq:third_term}).
Finally, we describe the complete super-resolution algorithm.

\subsection{Warping matrix construction} \label{sec:warp_matrices}

We define the set of the neighboring views ${\mathcal{N} _{k} ^{+}}$ in the term ${\mathcal{F} _{2}}$ in Eq.~(\ref{eq:second_term})
as containing only the four views ${\bm{U} _{k'}}$ adjacent to ${\bm{U} _{k}}$ in the light field:
\begin{equation*}
	\left\{ \bm{U} _{k'} : k' \in \mathcal{N} _{k} ^{+} \right\} \> = \> \left\{ \bm{U} _{s, t \pm 1}, \bm{U} _{s \pm 1, t} \right\}.
\end{equation*}
This choice reduces the number of the warping matrices but at the same time does not limit our problem formulation,
as the interlaced structure of the term ${\mathcal{F} _{2}}$ constrains together also those pairs of views that are not explicitly
constrained in ${\mathcal{F} _{2}}$.

The inner summation in Eq.~(\ref{eq:second_term}) considers the set of the four warping matrices ${ \{ \bm{F} _{k} ^{k'}: k' \in \mathcal{N} _{k} ^{+} \} }$
that warp the view ${\bm{U} _{k}}$ to each one of the four views ${\bm{U} _{k'}}$.
Conversely, but without loss of generality, in this section we consider the set of the four warping matrices ${ \{ \bm{F} _{k'} ^{k}: k' \in \mathcal{N} _{k} ^{+} \} }$
that warp each one of the four views ${\bm{U} _{k'}}$ to the view ${\bm{U} _{k}}$.
The warping matrix ${\bm{F} _{k'} ^{k}}$ is such that ${\bm{F} _{k'} ^{k} \bm{u} _{k'} \simeq \bm{u} _{k}}$.
In particular, the \mbox{$i$-th} row of the matrix ${\bm{F} _{k'} ^{k}}$ computes the pixel ${\bm{u} _{k} (i) = \bm{U} _{k} (x, y) = \bm{U} _{s, t} (x, y)}$
as a convex combination of those pixels around its projection on ${\bm{U} _{k'} = \bm{U} _{s', t'}}$.
Note that the convex combination is necessary, as the projection does not lie at integer spatial coordinates in general.
The exact position of the projection on ${\bm{U} _{s', t'}}$ is determined by the disparity value ${d _{x,y}}$ associated to the pixel ${\bm{U} _{s, t} (x, y)}$.
This is represented in Figure~\ref{fig:neighborhood}, which shows that the projections of ${\bm{U} _{s, t} (x, y)}$ on the four neighboring views
lie on the edges of a virtual square centered on the pixel ${\bm{U} _{s, t} (x, y)}$ and whose size depends on the disparity value ${d _{x,y}}$.
We estimate roughly the disparity value by finding a ${\delta \in \mathbb{N}}$ such that ${d _{x,y} \in [\delta, \delta + 1]}$.
In details, we first define a similarity score between the target pixel ${\bm{U} _{s, t} (x, y)}$ and a generic pixel ${\bm{U} _{s', t'} (x', y')}$ as follows:
\begin{equation} \label{eq:simil_score}
	w _{s', t'} \left( x', y' \right) \> = \> \exp{ \left( - \frac{ \| \mathcal{P} _{s, t} (x, y) - \mathcal{P} _{s', t'} (x', y') \| _{F} ^{2} }{\sigma ^{2}} \right) }
\end{equation}
where ${\mathcal{P} _{s, t} (x, y)}$ denotes a square patch centered at the pixel ${\bm{U} _{s, t} (x, y)}$, the operator ${\| \cdot \| _{F}}$
denotes the Frobenius norm, and $\sigma$ is a tunable constant. Then, we center a search window at ${\bm{U} _{s', t'} (x, y)}$
in each one of the four neighboring views, as represented in Figure~\ref{fig:neighborhood}.
In particular, we consider
\begin{itemize}
\item a ${1 \times W}$ pixel window for ${(s', t') = (s, t \pm 1)}$,
\item a ${W \times 1}$ pixel window for ${(s', t') = (s \pm 1, t)}$,
\end{itemize}
with ${W \in \mathbb{N}}$, and odd, defining the disparity range assumed for the whole light field,
i.e., ${d _{x, y} \in [- \lfloor W/2 \rfloor, \lfloor W/2 \rfloor] \> \forall (x, y)}$.
Finally, we introduce the following function that assigns a score to each possible value of $\delta$:
\begin{align} \label{eq:delta_cost_function}
\begin{split}
	\mathcal{S} \left( \delta \right) \>
	&= \> w _{s, t - 1} \left( x, y + \delta \right) \> + \> w _{s, t - 1} \left( x, y + \delta + 1 \right) \\
	&+ \> w _{s, t + 1} \left( x, y - \delta - 1 \right) \> + \> w _{s, t + 1} \left( x, y - \delta \right) \\
	&+ \> w _{s - 1, t} \left( x + \delta, y \right) \> + \> w _{s - 1, t} \left( x + \delta + 1, y \right) \\
	&+ \> w _{s + 1, t} \left( x - \delta - 1, y \right) \> + \> w _{s + 1, t} \left( x - \delta, y \right).
\end{split}
\end{align}
Note that each line of Eq.~(\ref{eq:delta_cost_function}) refers to a pair of adjacent pixels in one of the neighboring views.
We finally estimate $\delta$ as follows:
\begin{equation} \label{eq:delta_optimus}
	\delta ^{*} \> = \> \argmax _{\delta \in \left\{ - \lfloor W/2 \rfloor, \ldots, \lfloor W/2 \rfloor - 1 \right\} } \mathcal{S} \left( \delta \right).
\end{equation}
We can now fill the \mbox{$i$-th} row of the matrix ${\bm{F} _{k'} ^{k}}$ such that the pixel ${\bm{u} _{k} (i) = \bm{U} _{s, t} (x, y)}$
is computed as the convex combination of the two closest pixels to its projection on ${\bm{U} _{k'} = \bm{U} _{s', t'}}$,
namely the following two pixels:
\begin{align*}
\begin{split}
&\left\{ \bm{U} _{k'} \left( x, y + \delta \right), \bm{U} _{k'} \left( x, y + \delta + 1 \right) \right\} \> \text{for} \> (s', t') = (s, t - 1),\\
&\left\{ \bm{U} _{k'} \left( x, y - \delta - 1 \right), \bm{U} _{k'} \left( x, y - \delta \right) \right\} \> \text{for} \> (s', t') = (s, t + 1),\\
&\left\{ \bm{U} _{k'} \left( x + \delta, y \right), \bm{U} _{k'} \left( x + \delta + 1, y \right) \right\} \> \text{for} \> (s', t') = (s - 1, t),\\
&\left\{ \bm{U} _{k'} \left( x - \delta - 1, y \right), \bm{U} _{k'} \left( x - \delta, y \right) \right\} \> \text{for} \> (s', t') = (s + 1, t),
\end{split}
\end{align*}
which are indicated in green in Figure~\ref{fig:neighborhood}.
Once the two pixels involved in the convex combination at the \mbox{$i$-th} row of the matrix ${\bm{F} _{k'} ^{k}}$ are determined,
the \mbox{$i$-th} row can be constructed.
As an example, let us focus on the left neighboring view ${\bm{U} _{k'} = \bm{U} _{s, t-1}}$.
The two pixels involved in the convex combination at the \mbox{$i$-th} row of the matrix ${\bm{F} _{k'} ^{k}}$ are the following:
\begin{equation*}
	\{ \bm{u} _{k'} (j _{1}) = \bm{U} _{k'} \left( x, y + \delta \right), \bm{u} _{k'} (j _{2}) = \bm{U} _{k'} \left( x, y + \delta + 1 \right) \}.
\end{equation*}
We thus define the \mbox{$i$-th} row of the matrix ${\bm{F} _{k'} ^{k}}$ as follows:
\begin{equation*}
	\bm{F} _{k'} ^{k} \left( i, j \right) \> = \>
	\begin{cases}
		w _{s, t - 1} \left( x, y + \delta \right) / \> w \quad & \text{if \hspace{0.1cm}} j = j _{1} \\
		w _{s, t - 1} \left( x, y + \delta + 1 \right) / \> w \quad & \text{if \hspace{0.1cm}} j = j _{2} \\
		0 \quad & \text{otherwise} \\
	\end{cases}
\end{equation*}
with ${w \> \equiv \> w _{s, t - 1} (x, y + \delta) + w _{s, t - 1} (x, y + \delta + 1)}$.
In particular, each one of the two pixels in the convex combination has a weight that is proportional to its similarity to the
target pixel ${\bm{U} _{s, t} (x, y)}$.
For the remaining three neighboring views we proceed similarly.

We stress out that, for each pixel ${\bm{u} _{k} (i) = \bm{U} _{s, t} (x, y)}$, the outlined procedure fills the \mbox{$i$-th} row of
each one of the four matrices ${\bm{F} _{k'} ^{k}}$ with ${k' \in \mathcal{N} _{k} ^{+}}$.
In particular, as illustrated in Figure~\ref{fig:neighborhood}, the pair of pixels selected in each one of the four neighboring views encloses
one edge of the red square hosting the projections of the pixel ${\bm{U} _{s, t} (x, y)}$, therefore this procedure contributes to enforce
the light field structure in Eq.~(\ref{eq:twodim_stereo}).
Later on, we will refer to this particular structure as the \textit{square constraint}.

Finally, since occlusions are mostly handled by the regularizer ${\mathcal{F} _{3}}$, we use the corresponding masking matrix ${\bm{H} _{k'} ^{k}}$
in Eq.~(\ref{eq:second_term}) to handle only the trivial occlusions due to the image borders.

\subsection{Regularization graph construction} \label{sec:graph}

The effectiveness of the term $\mathcal{F} _{3}$ depends on the graph capability to capture the underlying structure of the light field.
Ideally, we would like to connect each pixel ${\bm{U} _{s, t} (x, y)}$ in the light field to its projections on the other views, as these
all share the same intensity value under the Lambertian assumption.
However, since the projections do not lie at integer spatial coordinates in general, we adopt a procedure similar to the warping matrix
construction and we aim at connecting the pixel ${\bm{U} _{s, t} (x, y)}$ to those pixels that are close to its projections on the other views.
We thus propose a three step approach to the computation of the graph adjacency matrix $\bm{W}$ in Eq.~(\ref{eq:graph_quad_form}).

\subsubsection{Edge weight computation}
We consider a view ${\bm{U} _{s, t}}$ and define its set of neighboring views ${\mathcal{N} _{k}}$ as follows:
\begin{equation*}
	\mathcal{N} _{k} \> \equiv \> \mathcal{N} _{k} ^{+} \> \cup \> \mathcal{N} _{k} ^{\times}
\end{equation*}
where we extend the neighborhood considered in the warping matrix construction with the four diagonal views.
In particular, ${\mathcal{N} _{k} ^{\times}}$ is defined as follows:
\begin{equation*}
	\left\{ \bm{U} _{k'} : k' \in \mathcal{N} _{k} ^{\times} \right\} \> = \> \left\{ \bm{U} _{s - 1, t \pm 1}, \bm{U} _{s + 1, t \pm 1} \right\}.
\end{equation*}
The full set of neighboring views is represented in Figure~\ref{fig:neighborhood}, with the views in ${\mathcal{N} _{k} ^{+}}$ in orange,
and those in ${\mathcal{N} _{k} ^{\times}}$ in green.
We then consider a pixel ${\bm{u} (i) = \bm{U} _{s, t} (x, y)}$ and define its edges toward one neighboring view ${\bm{U} _{k'} = \bm{U} _{s', t'}}$
with ${k' \in \mathcal{N} _{k}}$.
We center a search window at the pixel ${\bm{U} _{s', t'} (x, y)}$ and compute the following similarity score (weight) between the pixel
${\bm{U} _{s, t} (x, y) = \bm{u} (i)}$ and each pixel ${\bm{U} _{s', t'} (x', y')  = \bm{u} (j)}$ in the considered window:
\begin{equation} \label{eq:graph_weight}
	\bm{W} _{A} \left( i, j \right) \> = \> \exp{ \left( - \frac{ \| \mathcal{P} _{s, t} (x, y) - \mathcal{P} _{s', t'} (x', y') \| _{F} ^{2} }{\sigma ^{2}} \right) },
\end{equation}
with the notation already introduced in Section \ref{sec:warp_matrices}.
We repeat the procedure for each one of the eight neighboring views in ${\mathcal{N} _{k}}$, but we use differently shaped
windows at different views:
\begin{itemize}
\item a ${1 \times W}$ pixel window for ${(s', t') = (s, t \pm 1)}$,
\item a ${W \times 1}$ pixel window for ${(s', t') = (s \pm 1, t)}$,
\item a ${W \times W}$ pixel window otherwise.
\end{itemize}
This is illustrated in Figure~\ref{fig:neighborhood}.
The ${W \times W}$ pixel window is introduced for the diagonal views ${\bm{U} _{k} = \bm{U} _{s', t'}}$, with ${k' \in \mathcal{N} _{k} ^{\times}}$,
as the projection of the pixel ${\bm{U} _{s, t} (x, y)}$ on these views lies neither along row $x$, nor along column $y$.
Iterating the outlined procedure over each pixel ${\bm{u} (i)}$ in the light field leads to the construction of the adjacency matrix ${\bm{W} _{A}}$.
We regard ${\bm{W} _{A}}$ as the adjacency matrix of a \textit{directed} graph, with ${\bm{W} _{A} (i, j)}$ the weight of the edge from ${\bm{u} (i)}$
to ${\bm{u} (j)}$.

\subsubsection{Edge pruning}
We want to keep only the most important connections in the graph.
We thus perform a pruning of the edges leaving the pixel ${\bm{U} _{s, t} (x, y)}$ toward the eight neighboring views.
In particular, we keep only
\begin{itemize}
\item the two largest weight edges, for ${(s', t') = (s, t \pm 1)}$,
\item the two largest weight edges, for ${(s', t') = (s \pm 1, t)}$,
\item the four largest weight edges, otherwise.
\end{itemize}
For the diagonal neighboring views ${\bm{U} _{k'} = \bm{U} _{s', t'}}$, with ${k' \in \mathcal{N} _{k} ^{\times}}$, we allow
four weights rather than two as it is more difficult to detect those pixels that lie close to the projection of ${\bm{U} _{s, t} (x, y)}$.
We define ${\bm{W} _{B}}$ as the adjacency matrix after the pruning.

\subsubsection{Symmetric adjacency matrix}
We finally carry out the symmetrization of the matrix ${\bm{W} _{B}}$, and set ${\bm{W} \equiv \bm{W} _{B}}$ in Eq.~(\ref{eq:graph_quad_form}).
We adopt a simple approach for obtaining a symmetric matrix: we choose to preserve an edge between two vertexes $\bm{u} (i)$
and $\bm{u} (j)$ if and only if both entry $\bm{W} _{B} (i, j)$ and $\bm{W} _{B} (j, i)$ are non zero.
If this is the case, then $\bm{W} _{B} (i, j) = \bm{W} _{B} (j, i)$ necessarily holds true, and the weights are maintained.
This procedure mimics the well-known \textit{left-right disparity check} of stereo vision \cite{fua_parallel_1993}.

We finally note that the constructed graph can be used to build an alternative warping matrix to the one in Section~\ref{sec:warp_matrices}.
We recall that the matrix ${\bm{F} _{k'} ^{k}}$ is such that ${\bm{F} _{k'} ^{k} \bm{u} _{k'} \simeq \bm{u} _{k}}$.
In particular, the \mbox{$i$-th} row of this matrix is expected to compute the pixel ${\bm{u} _{k} (i) = \bm{U} _{s, t} (x, y)}$ as a
convex combination of those pixels around its projection on ${\bm{U} _{k'} = \bm{U} _{s', t'}}$.
We thus observe that the sub-matrix $\bm{W} _{S}$, obtained by extracting the rows ${(k - 1) N ^{2} + 1, \dots, k N ^{2}}$ and the columns
${(k' - 1) N ^{2}, \ldots, k' N ^{2}}$ from the adjacency matrix $\bm{W}$, represents a directed weighted graph with edges from the pixels of
the view ${\bm{U} _{k} = \bm{U} _{s, t}}$ (rows of the matrix) to the pixels of the view ${\bm{U} _{k'} = \bm{U} _{s', t'}}$ (columns of the matrix).
In this graph, the pixel ${\bm{u} _{k} (i) = \bm{U} _{s, t} (x, y)}$ is connected to a subset of pixels that lie close to its projections on
${\bm{U} _{k'} = \bm{U} _{s', t'}}$.
We thus normalize the rows of $\bm{W} _{S}$ such that they sum up to one, in order to implement a convex combination, and set
${\bm{F} _{k'} ^{k} \equiv \widetilde{\bm{W}} _{S}}$ with ${\widetilde{\bm{W}} _{S}}$ the normalized sub-matrix.
This alternative approach to the warping matrix construction does not take the light field structure explicitly into account,
but it represents a valid alternative when computational resources are limited, as it reuses the regularization graph.

\subsection{Optimization algorithm \label{ssec:opt_algo}}

We now have all the ingredients to solve the optimization problem in Eq.~(\ref{eq:objective_function}).
We observe that it corresponds to a quadratic problem.
We can first rewrite the first term, in Eq.~(\ref{eq:first_term}), as follows:
\begin{align} \label{eq:term_one}
\begin{split}
	\mathcal{F} _{1} \left( \bm{u} \right) \>
	&= \> \| \bm{A} \> \bm{u} \> - \> \bm{v} \| _{2} ^{2} \\
	&= \> \bm{u} ^{\top} \bm{A} ^{\top} \bm{A} \> \bm{u} \> - \> 2 \> \bm{v} ^{\top} \bm{A} \> \bm{u} \> + \> \bm{v} ^{\top} \bm{v}
\end{split}
\end{align}
with ${\bm{A} \equiv \bm{I} \otimes \bm{SB}}$, $\bm{I} \in \mathbb{R} ^{M ^{2}}$ the identity matrix, and $\otimes$ the Kronecker product.
For the second term, in Eq.~(\ref{eq:second_term}), we introduce the following matrices:
\begin{itemize}
\item ${\bm{H} _{k} \equiv \diag (\bm{H} _{k} ^{1}, \bm{H} _{k} ^{2}, \ldots, \bm{H} _{k} ^{M ^{2}})}$,
\item ${\bm{F} _{k} \equiv \bm{e} _{k} ^{\top} \otimes \left[(\bm{F} _{k} ^{1}) ^{\top} (\bm{F} _{k} ^{2}) ^{\top}
	\ldots (\bm{F} _{k} ^{M ^{2}}) ^{\top} \right] ^{\top}}$,
\end{itemize}
where $\diag$ denotes a block diagonal matrix, and ${\bm{e} _{k} \in \mathbb{R} ^{M ^{2}}}$ denotes the $k$-th vector
of the canonical basis, with ${\bm{e} _{k} (k) = 1}$ and zero elsewhere.
The matrices ${\bm{H} _{k} ^{k'}}$ and ${\bm{F} _{k} ^{k'}}$, originally defined only for ${k' \in \mathcal{N} _{k} ^{+}}$,
have been extended to the whole light field by assuming them to be zero for ${k' \notin \mathcal{N} _{k} ^{+}}$.
Finally, it is possible to remove the inner sum in Eq.~(\ref{eq:second_term}):
\begin{align} \label{eq:term_two}
	\mathcal{F} _{2} \left( \bm{u} \right) \>
	&= \> \sum _{k} \> \| \bm{H} _{k} \bm{A} \bm{F} _{k} \> \bm{u} \> - \> \bm{H} _{k} \bm{v} \| _{2} ^{2} \nonumber \\
	&= \> \sum _{k} \>
	\bm{u} ^{\top} \left( \bm{H} _{k} \bm{A} \bm{F} _{k} \right) ^{\top} \left( \bm{H} _{k} \bm{A} \bm{F} _{k} \right) \> \bm{u} \\
	&- \> \vphantom{\sum _{k}} 2 \> \left( \bm{H} _{k} \bm {v} \right) ^{\top} \left( \bm{H} _{k} \bm{A} \bm{F} _{k} \right) \bm{u} \>
	+ \> \left( \bm{H} _{k} \bm {v} \right) ^{\top} \left( \bm{H} _{k} \bm {v} \right). \nonumber
\end{align}
Replacing Eq.~(\ref{eq:term_one}) and Eq.~(\ref{eq:term_two}) in Eq.~(\ref{eq:objective_function}) permits to rewrite the objective function
$\mathcal{F} (\bm{u})$ in a quadratic form:
\begin{align} \label{eq:objective_standard}
	\bm{u} ^{*} \> &\in \> \argmin _{\bm{u}} \>\>
	\underbrace{
	\frac{1}{2} \> \bm{u} ^{\top} \bm{P} \> \bm{u} \> + \> \bm{q} ^{\top} \bm{u} \> + \> r
	} _{\mathcal{F} \left( \bm{u} \right)}
\end{align}
with
\begin{align*}
	\bm{P} \> &\equiv \> 2 \left(
		\bm{A} ^{\top} \bm{A}
		+ \lambda _{2} \sum _{k} \left( \bm{H} _{k} \bm{A} \bm{F} _{k} \right) ^{\top} \left( \bm{H} _{k} \bm{A} \bm{F} _{k} \right)
		+ \lambda _{3} \bm{L} \right) \\
	\bm{q} \> &\equiv \>
		-2 \left(
		\bm{A} ^{\top} \>
		+ \> \lambda _{2} \sum _{k} \left( \left( \bm{H} _{k} \bm{A} \bm{F} _{k} \right) ^ {\top} \bm{H} _{k} \right)
		\right) \bm{v} \\
	r \> &\equiv \>
		\bm{v} ^{\top} \left( \bm{I} \> + \> \lambda _{2} \sum _{k} \left( \bm{H} _{k} ^{\top} \bm{H} _{k} \right) \right) \bm{v}.
\end{align*}

We observe that, in general, the matrix $\bm{P}$ is positive semi-definite, therefore it is not possible to solve Eq.~(\ref{eq:objective_standard})
just by employing the \textit{Conjugate Gradient (CG)} method on the linear system ${\nabla \mathcal{F} (\bm{u}) = \bm{P} \bm{u} - \bm{q} = 0}$.
We thus choose to adopt the \textit{Proximal Point Algorithm (PPA)}, which iteratively solves Eq.~(\ref{eq:objective_standard}) using the following
update rule:
\begin{align}
	\bm{u} ^{(i + 1)} \>
	&= \> \prox _{\beta \mathcal{F}} \left( \bm{u} ^{ \left( i \right) } \right) \\
	&= \> \argmin _{\bm{u}} \>\>
		\mathcal{F} \left( \bm{u} \right) \> + \> \frac{1}{2 \beta} \| \bm{u} - \bm{u} ^{ \left( i \right) } \| _{2} ^{2} \nonumber \\
	&= \> \argmin _{\bm{u}} \>\>
		\underbrace{
		\frac{1}{2} \> \bm{u} ^{\top} \left( \bm{P} + \frac{\bm{I}}{\beta} \right) \> \bm{u} \>
		+ \> \left( \bm{q} - \frac{\bm{u} ^{ \left( i \right) }}{\beta} \right) ^{\top} \bm{u}
		} _{\mathcal{T} \left( \bm{u} \right)}. \nonumber
\end{align}
The matrix ${\bm{P} + (1 / \beta) \bm{I}}$ is positive definite for every $\beta > 0$, hence we can now use the CG method
to solve the linear system ${\nabla{\mathcal{T} (\bm{u})} = 0}$.
The full Graph-Based super-resolution algorithm is summarized in Algorithm~\ref{algo:full_super_algo}.
We observe that both the construction of the warping matrices and of the graph requires the high resolution light field to be available.
In order to bypass this causality problem, a fast and rough high resolution estimation of the light field is computed, e.g., via bilinear interpolation,
at the bootstrap phase.
Then, at each new iteration, the warping matrices and the graph can be reconstructed on the new available estimate of the high resolution
light field, and a new estimate can be obtained by solving the problem in Eq.~(\ref{eq:objective_standard}).

\begin{algorithm}[t]
	\caption{Graph-Based Light Field Super-Resolution}
	\label{algo:full_super_algo}
	\begin{algorithmic}[1]
	\Input ${\bm{v} = [ \bm{v} _{1}, \ldots, \bm{v} _{M ^{2}} ]}$, ${\alpha \in \mathbb{N}}$, ${\beta > 0}$, $iter$.
	\Output ${\bm{u} = [ \bm{u} _{1}, \ldots, \bm{u} _{M ^{2}} ]}$.
	\State ${\bm{u} \gets}$ bilinear interp. of ${\bm{v} _{k}}$ by $\alpha$, ${\forall k = 1, \ldots, M ^{2}}$;
	\For{$i = 1 \> \> \textbf{to} \> \> iter$}
		\State build the graph adjacency matrix $\bm{W}$ on $\bm{u}$;
		\State build the matrices $\bm{F} _{k}$ on $\bm{u}$, ${\forall k = 1, \ldots, M ^{2}}$;
		\State update the matrix $\bm{P}$;
		\State update the vector $\bm{q}$;
		\State ${\bm{z} \gets \bm{u}}$; \Comment Initialize CG
		\While{convergence is reached}
			\State ${\bm{z} \gets}$ CG${( \bm{P} + (\bm{I} / \beta), (\bm{z} / \beta) - \bm{q})}$;
		\EndWhile
		\State $\bm{u} \gets \bm{z}$; \Comment Update $\bm{u}$
	\EndFor
	\State \textbf{return} $\bm{u}$;
	\end{algorithmic}
\end{algorithm}
%

\section{Computational Complexity} \label{sec:complexity}

In this section we provide an estimate of the computational complexity of our super-resolution algorithm proposed
in Section~\ref{sec:super_res_algo}.
This is comprised of three main steps:
the construction of the graph adjacency matrix,
the construction of the warping matrices,
and the solution of the optimization problem in Eq.~(\ref{eq:objective_standard}).
We analyze each one of these steps separately.

In the graph construction step, the weights from each view to the eight neighboring ones are computed.
Using the method in \cite{darbon_fast_2008}, the computation of all the weights from one view to the eight neighboring ones can
be made independent of the size of the square patch $\mathcal{P}$ and computed in ${O(N ^{2} W ^{2})}$ operations,
where ${N ^{2}}$ is the number of pixels per view and ${W ^{2}}$ is the maximum number of pixels in a search window.
This balance takes into account also the operations required by the selection of the highest weights,
which is necessary to define the graph edges.
Repeating this procedure for all the ${M ^{2}}$ views in the light field leads to a complexity ${O(M ^{2} N ^{2} W ^{2})}$,
or equivalently ${O(M ^{2} N ^{2} \alpha ^{2})}$, as the disparity in the high resolution views grows with the super-resolution
factor $\alpha$, and it is therefore reasonable to define the size $W$ of the search window as a multiple of $\alpha$.

The construction of the warping matrices relies on the previously computed weights, therefore the complexity of this step depends
exclusively on the estimation of the parameter $\delta$ in Eq.~(\ref{eq:delta_optimus}).
The computation of $\delta$ for each pixel in a view requires ${O(N ^{2} W)}$ operations, where $W$ is no longer squared
because only 1D search windows are considered at this step.
The computation of $\delta$ for all the views in the light field leads to a complexity ${O(M ^{2} N ^{2} W)}$,
or equivalently ${O(M ^{2} N ^{2} \alpha)}$.

Finally, the optimization problem in Eq.~(\ref{eq:objective_standard}) is solved via PPA, whose iterations consist in a call to the CG method
(cf. steps 8-10 in Algorithm~\ref{algo:full_super_algo}).
Each internal iteration of the CG method is dominated by a matrix-vector multiplication with the ${M ^{2} N ^{2} \times M ^{2} N ^{2}}$
matrix ${\bm{P} + (\bm{I} / \beta)}$.
However, it is straightforward to observe that the matrix ${\bm{P} + (\bm{I} / \beta)}$ is very sparse with ${O(M ^{2} N ^{2} \alpha ^{4})}$
non zeros entries, where we assume the size of the blurring kernel to be ${\alpha \times \alpha}$ pixels, as in our tests
in Section~\ref{sec:experiments}.
It follows that the matrix-vector multiplication within each CG internal iteration requires only ${O(M ^{2} N ^{2} \alpha ^{4})}$ operations.
The complexity of the overall optimization step depends on the number of iterations of PPA, and on the number of internal iterations
performed by each instance of CG.
Although we do not provide an analysis of the convergence rate of our optimization algorithm, in our tests we empirically observe the following:
regardless of the number of pixels in the high resolution light field, in general PPA converges after ${30}$ iterations
(each one consisting in a call to CG) while each instance of CG typically converges in only $9$ iterations.
Therefore, assuming the global number of iterations of CG to be independent of the light field size,
we approximate the complexity of the optimization step with ${O(M ^{2} N ^{2} \alpha ^{4})}$.

The global complexity of our super-resolution algorithm can finally be approximated with ${O(M ^{2} N ^{2} \alpha ^{4})}$,
which is linear in the number of pixels of the high resolution light field, hence it represents a reasonable complexity.
Moreover, we observe that the graph and warping matrix construction steps can be highly parallelized.
Although this feature would not affect the algorithm computational complexity, in practice it could lead to a significative speed up.
Finally, compared to the light field super-resolution method in \cite{wanner_spatial_2012}, which employs TV regularization,
our algorithm turns super-resolution into a simpler (quadratic) optimization problem, and differently from the learning-based
light field super-resolution method in \cite{mitra_light_2012} it does not require any time demanding training.

\section{Experiments} \label{sec:experiments}

\subsection{Experimental settings} \label{subsec:exp_settings}

We now provide extensive experiments to analyze the performance of our algorithm.
We compare it to the two super-resolution algorithms that, up to our knowledge, are the only ones developed explicitly for light field data,
and that we already introduced in Section~\ref{sec:related_work}: Wanner and Goldluecke \cite{wanner_spatial_2012},
and Mitra and Veeraraghavan \cite{mitra_light_2012}.
We also compare our algorithm to the CNN-based super-resolution algorithm in \cite{dong_learning_2014}, which represent
the state-of-the-art for single-frame super-resolution.
Up to the authors knowledge, a multi-frame super-resolution algorithm based on CNNs has not been presented yet.

We test our algorithm on two public datasets: the \textit{HCI light field dataset} \cite{wanner_datasets_2013} and the
\textit{(New) Stanford light field dataset} \cite{stanford_dataset}.
The HCI dataset comprises twelve light fields, each one characterized by a ${9 \times 9}$ array of views.
Seven light fields have been artificially generated with a 3D creation suite, while the remaining five have been acquired with a traditional
SLR camera mounted on a motorized gantry, that permits to move the camera precisely and emulate a camera array with an arbitrary baseline.
The HCI dataset is meant to represent the data from a light field camera, where both the baseline distance $b$ between adjacent views
and the disparity range are typically very small.
In particular, in the HCI dataset the disparity range is within ${[-3,3]}$ pixels.
Differently, the Stanford dataset contains light fields whose view baseline and disparity range can be much larger.
For this reason, the Stanford dataset does not closely resemble the typical data from a light field camera.
However, we include the Stanford dataset in our experiments in order to evaluate the robustness of light field super-resolution methods
to larger disparity ranges, possibly exceeding the assumed one.
The Stanford light fields have all been acquired with a gantry, and they are characterized by a ${17 \times 17}$ array of views.
Similarly to \cite{wanner_spatial_2012} and \cite{mitra_light_2012}, in our experiments we crop the light fields to a ${5 \times 5}$ array of views,
i.e., we choose ${M = 5}$.

In our experiments, we first create the low resolution version of the test light field from the datasets above.
The spatial resolution of the test light field $\bm{U}$ is decreased by a factor ${\alpha \in \mathbb{N}}$ by applying the blurring and
sampling matrix $\bm{SB}$ of Eq.~(\ref{eq:hr2lr}) to each color channel of each view.
In order to match the assumptions of the other light field super-resolution frameworks involved in the comparison \cite{wanner_spatial_2012}
\cite{mitra_light_2012}, and without loss of generality, the blur kernel implemented by the matrix $\bm{B}$ is set to an ${\alpha \times \alpha}$ box filter,
and the matrix $\bm{S}$ performs a regular sampling.
Then the obtained low resolution light field $\bm{V}$ is brought back to the original spatial resolution by the super-resolution algorithms under study.
This approach guarantees that the final spatial resolution of the test light field is exactly its original one, regardless of $\alpha$. 

In our framework, the low resolution light field $\bm{V}$ is divided into possibly overlapping sub-light-fields and each one is
reconstructed separately.
Formally, a sub-light-field is obtained by fixing a spatial coordinate ${(x, y)}$ and then extracting an ${N' \times N'}$ patch with the top left
pixel at ${(x, y)}$ from each view ${\bm{V} _{s, t}}$.
The result is an ${N' \times N' \times M \times M}$ light field, with ${N' < (N / \alpha)}$.
Once all the sub-light-fields are super resolved, different estimates of the same pixel produced by the possible overlap are merged.
We set ${N' = 100}$ and $70$ for ${\alpha = 2}$ and $3$, respectively.
This choice leads to a high resolution sub-light-field with a spatial resolution that is approximatively $200 \times 200$ pixels.
Finally, only the luminance channel of the low resolution light field is super resolved using our method, as the two chrominance channels
can be easily up-sampled via bilinear interpolation due to their low frequency nature.

\begin{table*}
	\centering
	\caption{hci dataset - psnr mean and variance for the super-resolution factor $\alpha = 2$}
	\label{tab:hci2}
	\begin{tabular}{| l | c | c | c | c | c | c |}
	\hline
	& Bilinear & \cite{wanner_spatial_2012} & \cite{mitra_light_2012} & \cite{dong_learning_2014} & GB-DR & GB-SQ \\
	\hline
	\hline
	buddha & 35.22 $\pm$ 0.00 & 38.22 $\pm$ 0.11 & \textbf{39.12} $\pm$ 0.62 & 37.73 $\pm$ 0.03 & 38.59 $\pm$ 0.08 & 39.00 $\pm$ 0.14 \\
	\hline
	buddha2 & 30.97 $\pm$ 0.00 & 33.01 $\pm$ 0.11 & 33.63 $\pm$ 0.22 & 33.67 $\pm$ 0.00 & 34.17 $\pm$ 0.01 & \textbf{34.41} $\pm$ 0.02 \\
	\hline
	couple & 25.52 $\pm$ 0.00 & 26.22 $\pm$ 1.61 & 31.83 $\pm$ 2.80 & 28.56 $\pm$ 0.00 & 32.79 $\pm$ 0.17 & \textbf{33.51} $\pm$ 0.25 \\
	\hline
	cube & 26.06 $\pm$ 0.00 & 26.40 $\pm$ 1.90 & 30.99 $\pm$ 3.02 & 28.81 $\pm$ 0.00 & 32.60 $\pm$ 0.23 & \textbf{33.28} $\pm$ 0.35 \\
	\hline
	horses & 26.37 $\pm$ 0.00 & 29.14 $\pm$ 0.63 & \textbf{33.13} $\pm$ 0.72 & 27.80 $\pm$ 0.00 & 30.99 $\pm$ 0.05 & 32.62 $\pm$ 0.12 \\
	\hline
	maria & 32.84 $\pm$ 0.00 & 35.60 $\pm$ 0.33 & 37.03 $\pm$ 0.44 & 35.50 $\pm$ 0.00 & 37.19 $\pm$ 0.03 & \textbf{37.25} $\pm$ 0.02 \\
	\hline
	medieval & 30.07 $\pm$ 0.00 & 30.53 $\pm$ 0.67 & 33.34 $\pm$ 0.71 & 31.23 $\pm$ 0.00 & 33.23 $\pm$ 0.03 & \textbf{33.45} $\pm$ 0.02 \\
	\hline
	mona & 35.11 $\pm$ 0.00 & 37.54 $\pm$ 0.64 & 38.32 $\pm$ 1.14 & 39.07 $\pm$ 0.00 & 39.30 $\pm$ 0.04 & \textbf{39.37} $\pm$ 0.05 \\
	\hline
	papillon & 36.19 $\pm$ 0.00 & 39.91 $\pm$ 0.15 & 40.59 $\pm$ 0.89 & 39.88 $\pm$ 0.00 & \textbf{40.94} $\pm$ 0.06 & 40.70 $\pm$ 0.04 \\
	\hline
	pyramide & 26.49 $\pm$ 0.00 & 26.73 $\pm$ 1.42 & 33.35 $\pm$ 4.06 & 29.69 $\pm$ 0.00 & 34.63 $\pm$ 0.34 & \textbf{35.41} $\pm$ 0.67 \\
	\hline
	statue & 26.32 $\pm$ 0.00 & 26.15 $\pm$ 2.15 & 32.95 $\pm$ 4.67 & 29.65 $\pm$ 0.00 & 34.81 $\pm$ 0.38 & \textbf{35.61} $\pm$ 0.73 \\
	\hline
	stillLife & 25.28 $\pm$ 0.00 & 25.58 $\pm$ 1.41 & 28.84 $\pm$ 0.82 & 27.27 $\pm$ 0.00 & 30.80 $\pm$ 0.07 & \textbf{30.98} $\pm$ 0.05 \\
	\hline
	\end{tabular}
\end{table*}
\begin{table*}
	\centering
	\caption{stanford dataset - psnr mean and variance for the super-resolution factor $\alpha = 2$}
	\label{tab:stanford2}
	\begin{tabular}{| l | c | c | c | c | c | c |}
	\hline
	& Bilinear & \cite{wanner_spatial_2012} & \cite{mitra_light_2012} & \cite{dong_learning_2014} & GB-DR & GB-SQ \\
	\hline
	\hline
	amethyst & 35.59 $\pm$ 0.01 & 24.18 $\pm$ 0.20 & 36.08 $\pm$ 4.12 & 38.81 $\pm$ 0.01 & \textbf{40.30} $\pm$ 0.11 & 39.19 $\pm$ 0.25 \\
	\hline
	beans & 47.92 $\pm$ 0.01 & 23.28 $\pm$ 0.53 & 36.02 $\pm$ 7.38 & \textbf{52.01} $\pm$ 0.01 & 50.20 $\pm$ 0.16 & 48.41 $\pm$ 1.18 \\
	\hline
	bracelet & 33.02 $\pm$ 0.00 & 18.98 $\pm$ 0.22 & 19.91 $\pm$ 3.86 & 38.05 $\pm$ 0.00 & \textbf{39.10} $\pm$ 0.43 & 28.27 $\pm$ 2.84 \\
	\hline
	bulldozer & 34.94 $\pm$ 0.01 & 22.82 $\pm$ 0.09 & 32.05 $\pm$ 3.73 & \textbf{39.76} $\pm$ 0.03 & 37.27 $\pm$ 0.33 & 35.96 $\pm$ 0.43 \\
	\hline
	bunny & 42.44 $\pm$ 0.01 & 26.22 $\pm$ 1.15 & 40.66 $\pm$ 6.69 & \textbf{47.16} $\pm$ 0.01 & 46.96 $\pm$ 0.06 & 47.01 $\pm$ 0.06 \\
	\hline
	cards & 29.50 $\pm$ 0.00 & 19.38 $\pm$ 0.22 & 28.46 $\pm$ 5.68 & 33.77 $\pm$ 0.00 & \textbf{36.54} $\pm$ 0.20 & 36.52 $\pm$ 0.20 \\
	\hline
	chess & 36.36 $\pm$ 0.00 & 21.77 $\pm$ 0.27 & 34.74 $\pm$ 5.85 & 40.75 $\pm$ 0.00 & \textbf{42.04} $\pm$ 0.08 & 41.86 $\pm$ 0.07 \\
	\hline
	eucalyptus & 34.09 $\pm$ 0.00 & 25.04 $\pm$ 0.28 & 34.90 $\pm$ 3.50 & 36.69 $\pm$ 0.00 & 39.07 $\pm$ 0.12 & \textbf{39.09} $\pm$ 0.08 \\
	\hline
	knights & 34.31 $\pm$ 0.04 & 21.14 $\pm$ 0.24 & 29.33 $\pm$ 4.77 & 38.37 $\pm$ 0.06 & \textbf{39.68} $\pm$ 0.27 & 37.23 $\pm$ 0.40 \\
	\hline
	treasure & 30.83 $\pm$ 0.00 & 22.81 $\pm$ 0.15 & 30.52 $\pm$ 2.93 & 34.16 $\pm$ 0.00 & \textbf{37.68} $\pm$ 0.26 & 37.51 $\pm$ 0.15 \\
	\hline
	truck & 36.26 $\pm$ 0.04 & 25.77 $\pm$ 0.08 & 37.52 $\pm$ 4.60 & 39.11 $\pm$ 0.09 & 41.09 $\pm$ 0.14 & \textbf{41.57} $\pm$ 0.15 \\
	\hline
	\end{tabular}
\end{table*}
\begin{table*}
	\centering
	\caption{hci dataset - psnr mean and variance for the super-resolution factor $\alpha = 3$}
	\label{tab:hci3}
	\begin{tabular}{| l | c | c | c | c | c | c |}
	\hline
	& Bilinear & \cite{wanner_spatial_2012} & \cite{mitra_light_2012} & \cite{dong_learning_2014} & GB-DR & GB-SQ \\
	\hline
	\hline
	buddha & 32.58 $\pm$ 0.01 & 34.92 $\pm$ 0.63 & 35.36 $\pm$ 0.34 & 34.62 $\pm$ 0.01 & \textbf{35.42} $\pm$ 0.02 & 35.40 $\pm$ 0.02 \\
	\hline
	buddha2 & 28.14 $\pm$ 0.00 & 29.96 $\pm$ 0.07 & 30.29 $\pm$ 0.10 & 30.23 $\pm$ 0.00 & \textbf{30.52} $\pm$ 0.00 & 30.43 $\pm$ 0.00 \\
	\hline
	couple & 22.62 $\pm$ 0.00 & 23.02 $\pm$ 1.56 & \textbf{27.43} $\pm$ 1.16 & 24.01 $\pm$ 0.00 & 26.65 $\pm$ 0.01 & 26.95 $\pm$ 0.00 \\
	\hline
	cube & 23.25 $\pm$ 0.00 & 22.47 $\pm$ 2.65 & 26.48 $\pm$ 1.16 & 24.58 $\pm$ 0.00 & 27.23 $\pm$ 0.01 & \textbf{27.39} $\pm$ 0.00 \\
	\hline
	horses & 24.35 $\pm$ 0.00 & 24.45 $\pm$ 1.27 & \textbf{29.90} $\pm$ 0.55 & 24.73 $\pm$ 0.00 & 25.53 $\pm$ 0.00 & 26.41 $\pm$ 0.01 \\
	\hline
	maria & 30.02 $\pm$ 0.00 & 30.64 $\pm$ 0.87 & 33.36 $\pm$ 0.37 & 31.55 $\pm$ 0.00 & \textbf{33.48} $\pm$ 0.01 & 33.12 $\pm$ 0.01 \\
	\hline
	medieval & 28.29 $\pm$ 0.00 & 28.54 $\pm$ 0.37 & \textbf{29.78} $\pm$ 0.50 & 28.57 $\pm$ 0.00 & 29.23 $\pm$ 0.00 & 29.54 $\pm$ 0.01 \\
	\hline
	mona & 32.05 $\pm$ 0.00 & 33.42 $\pm$ 0.52 & 33.31 $\pm$ 0.40 & \textbf{34.82} $\pm$ 0.00 & 34.66 $\pm$ 0.01 & 34.47 $\pm$ 0.01 \\
	\hline
	papillon & 33.66 $\pm$ 0.00 & \textbf{36.76} $\pm$ 0.13 & 36.13 $\pm$ 0.48 & 36.56 $\pm$ 0.00 & 36.44 $\pm$ 0.01 & 36.18 $\pm$ 0.01 \\
	\hline
	pyramide & 23.39 $\pm$ 0.00 & 23.60 $\pm$ 2.72 & \textbf{29.13} $\pm$ 1.86 & 24.84 $\pm$ 0.00 & 28.34 $\pm$ 0.01 & 28.48 $\pm$ 0.00 \\
	\hline
	statue & 23.21 $\pm$ 0.00 & 22.97 $\pm$ 3.63 & \textbf{28.93} $\pm$ 2.03 & 24.72 $\pm$ 0.00 & 28.21 $\pm$ 0.01 & 28.38 $\pm$ 0.00 \\
	\hline
	stillLife & 23.28 $\pm$ 0.00 & 23.62 $\pm$ 1.64 & \textbf{27.23} $\pm$ 0.49 & 23.83 $\pm$ 0.00 & 24.99 $\pm$ 0.00 & 25.54 $\pm$ 0.00 \\
	\hline
	\end{tabular}
\end{table*}
\begin{table*}
	\centering
	\caption{stanford dataset - psnr mean and variance for the super-resolution factor $\alpha = 3$}
	\label{tab:stanford3}
	\begin{tabular}{| l | c | c | c | c | c | c |}
	\hline
	& Bilinear & \cite{wanner_spatial_2012} & \cite{mitra_light_2012} & \cite{dong_learning_2014} & GB-DR & GB-SQ \\
	\hline
	\hline
	amethyst & 32.69 $\pm$ 0.01 & 21.94 $\pm$ 0.30 & 31.47 $\pm$ 1.07 & 34.79 $\pm$ 0.01 & \textbf{35.97} $\pm$ 0.03 & 35.63 $\pm$ 0.02 \\
	\hline
	beans & 43.81 $\pm$ 0.01 & 22.66 $\pm$ 0.26 & 31.25 $\pm$ 1.85 & 47.38 $\pm$ 0.02 & \textbf{48.67} $\pm$ 0.12 & 47.28 $\pm$ 0.43 \\
	\hline
	bracelet & 29.06 $\pm$ 0.00 & 17.37 $\pm$ 0.22 & 15.83 $\pm$ 0.32 & 31.96 $\pm$ 0.00 & \textbf{35.23} $\pm$ 0.06 & 30.46 $\pm$ 1.08 \\
	\hline
	bulldozer & 31.88 $\pm$ 0.00 & 21.85 $\pm$ 0.11 & 26.21 $\pm$ 0.85 & \textbf{35.48} $\pm$ 0.01 & 35.44 $\pm$ 0.05 & 35.31 $\pm$ 0.04 \\
	\hline
	bunny & 39.03 $\pm$ 0.00 & 23.40 $\pm$ 1.22 & 35.88 $\pm$ 1.82 & 43.62 $\pm$ 0.01 & \textbf{43.73} $\pm$ 0.05 & 43.54 $\pm$ 0.05 \\
	\hline
	cards & 26.13 $\pm$ 0.00 & 17.77 $\pm$ 0.33 & 25.22 $\pm$ 1.40 & 28.34 $\pm$ 0.00 & 31.03 $\pm$ 0.02 & \textbf{31.49} $\pm$ 0.03 \\
	\hline
	chess & 33.11 $\pm$ 0.00 & 20.56 $\pm$ 0.24 & 31.19 $\pm$ 1.96 & 35.76 $\pm$ 0.00 & \textbf{36.87} $\pm$ 0.04 & 36.76 $\pm$ 0.03 \\
	\hline
	eucalyptus & 31.71 $\pm$ 0.00 & 23.38 $\pm$ 0.17 & 32.23 $\pm$ 1.61 & 33.03 $\pm$ 0.00 & 34.51 $\pm$ 0.01 & \textbf{34.80} $\pm$ 0.01 \\
	\hline
	knights & 31.31 $\pm$ 0.02 & 19.36 $\pm$ 0.07 & 25.55 $\pm$ 1.40 & 34.38 $\pm$ 0.05 & \textbf{35.37} $\pm$ 0.06 & 35.21 $\pm$ 0.05 \\
	\hline
	treasure & 27.98 $\pm$ 0.00 & 21.45 $\pm$ 0.14 & 27.86 $\pm$ 0.89 & 29.58 $\pm$ 0.00 & \textbf{31.37} $\pm$ 0.01 & 31.21 $\pm$ 0.01 \\
	\hline
	truck & 33.52 $\pm$ 0.02 & 23.27 $\pm$ 0.05 & 33.04 $\pm$ 1.66 & 35.45 $\pm$ 0.04 & 36.67 $\pm$ 0.05 & \textbf{36.97} $\pm$ 0.05 \\
	\hline
	\end{tabular}
\end{table*}

In our experiments we consider two variants of our Graph-Based super-resolution algorithm (GB).
The first is \textit{GB-SQ}, the main variant, which employs the warping matrix construction based on the square constraint (SQ) and is presented
in Section~\ref{sec:warp_matrices}.
The second is \textit{GB-DR}, the variant employing the warping matrices extracted directly (DR) from the graph and introduced
at the end of Section~\ref{sec:graph}.
In the warping matrix construction in Eq.~(\ref{eq:simil_score}), as well as in the graph construction in Eq.~(\ref{eq:graph_weight}),
we empirically set the size of the patch $\mathcal{P}$ to ${7 \times 7}$ pixels and ${\sigma = 0.7229}$.
For the search window size, we set ${W = 13}$ pixels.
This choice is equivalent to consider a disparity range of ${[-6, 6]}$ pixels at high resolution.
Note that this range happens to be loose for the HCI dataset, whose disparity range is within ${[-3, 3]}$.
Choosing exactly the correct disparity range for each light field could both improve the reconstruction by avoiding possible wrong correspondences
in the graph and warping matrices, and decrease the computation time.
On the other hand, for some light fields in the Stanford dataset, the chosen disparity range may become too small, thus preventing
the possibility to capture the correct correspondences.
In practice, the disparity range is not always available, hence the value ${[-6, 6]}$ happens to be a fair choice considering that our super-resolution
framework targets data from light field cameras, i.e., data with a typically small disparity range.
Finally, without any fine tuning on the considered datasets, we empirically set ${\lambda _{2} = 0.2}$ and ${\lambda _{3} = 0.0055}$
in the objective function in Eq.~$(\ref{eq:objective_function})$ and perform just two iterations of the full Algorithm \ref{algo:full_super_algo},
as we experimentally found them to be sufficient.

\begin{figure*}
	\centering
	\subfloat[LR]{%
		\includegraphics[height=3cm]{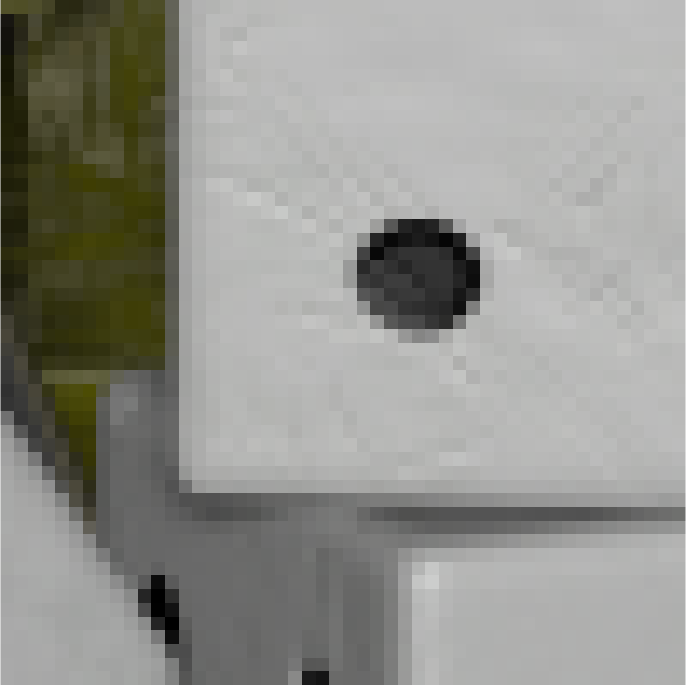}
	}
	~
	\subfloat[Bilinear]{%
		\includegraphics[height=3cm]{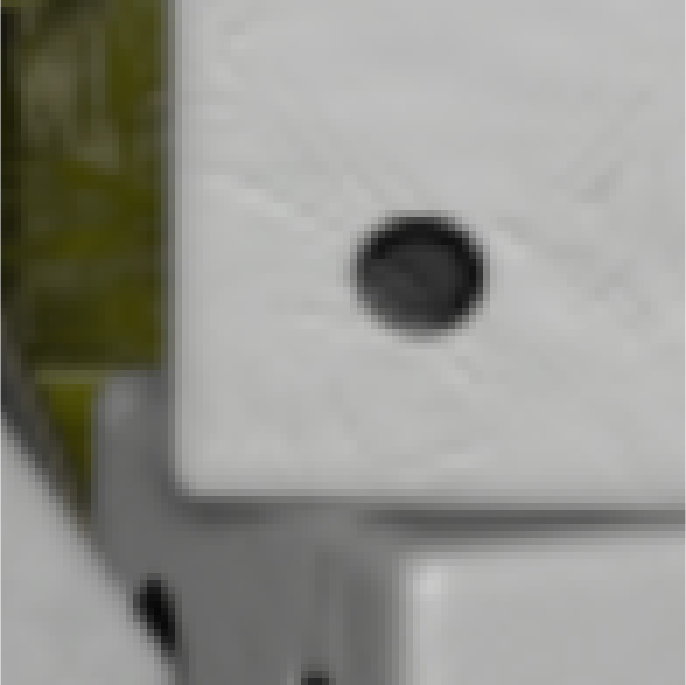}
		\label{fig:hci2_buddhaBIL}
	}
	~
	\subfloat[ \cite{wanner_spatial_2012} ]{%
		\includegraphics[height=3cm]{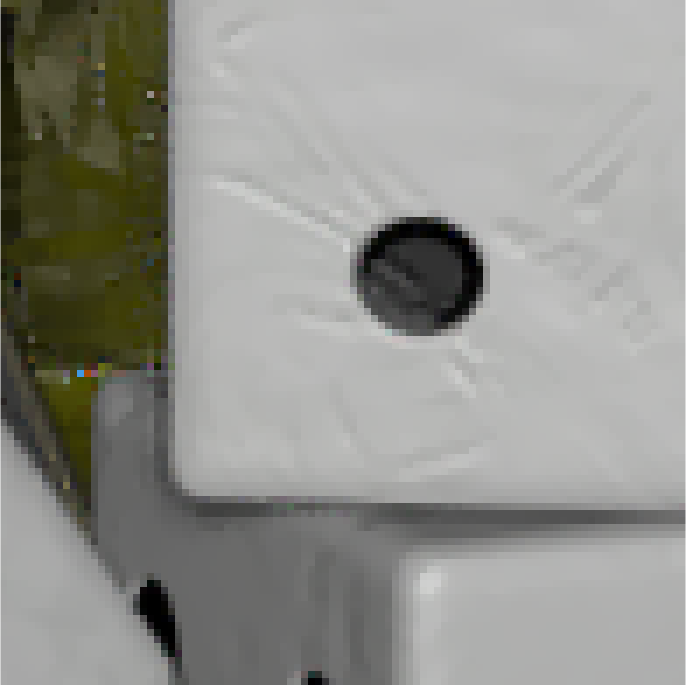}
		\label{fig:hci2_buddhaCOCO}
	}
	~
	\subfloat[ \cite{mitra_light_2012} ]{%
		\includegraphics[height=3cm]{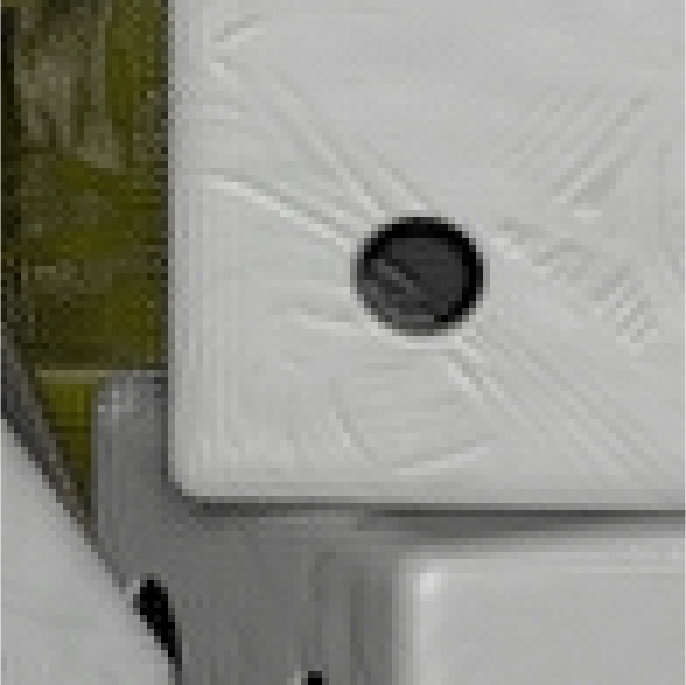}
		\label{fig:hci2_buddhaGMM}
	}
	\vspace{0.1cm}
	\subfloat[ \cite{dong_learning_2014} ]{%
		\includegraphics[height=3cm]{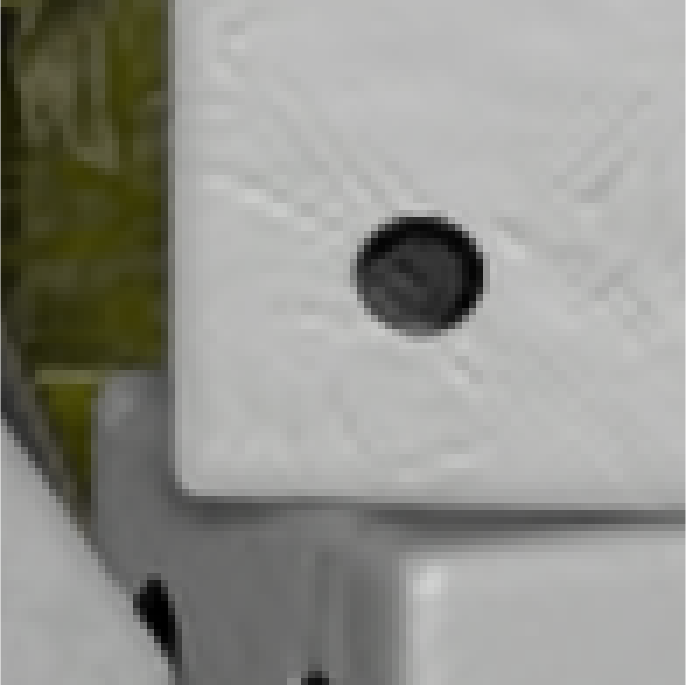}
		\label{fig:hci2_buddhaSRCNN}
	}
	~
	\subfloat[GB-DR]{%
		\includegraphics[height=3cm]{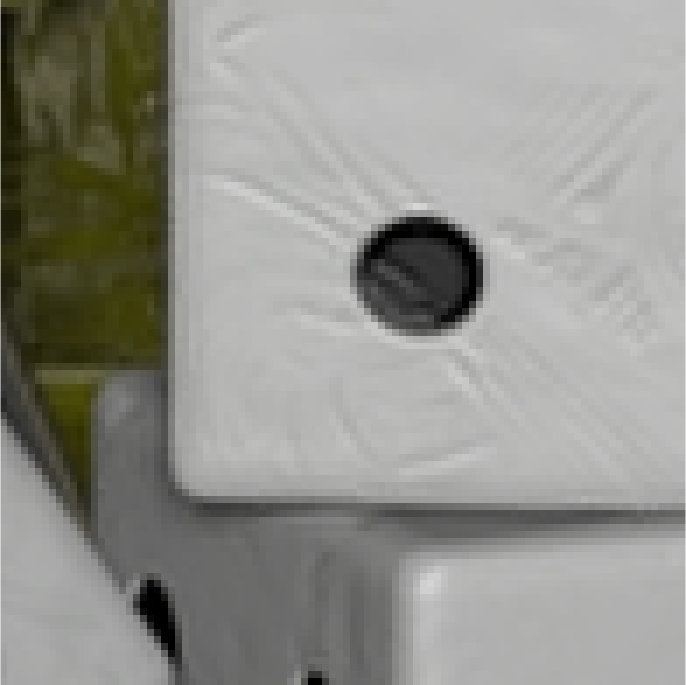}
		\label{fig:hci2_buddhaGBDR}
	}
	~
	\subfloat[GB-SQ]{%
		\includegraphics[height=3cm]{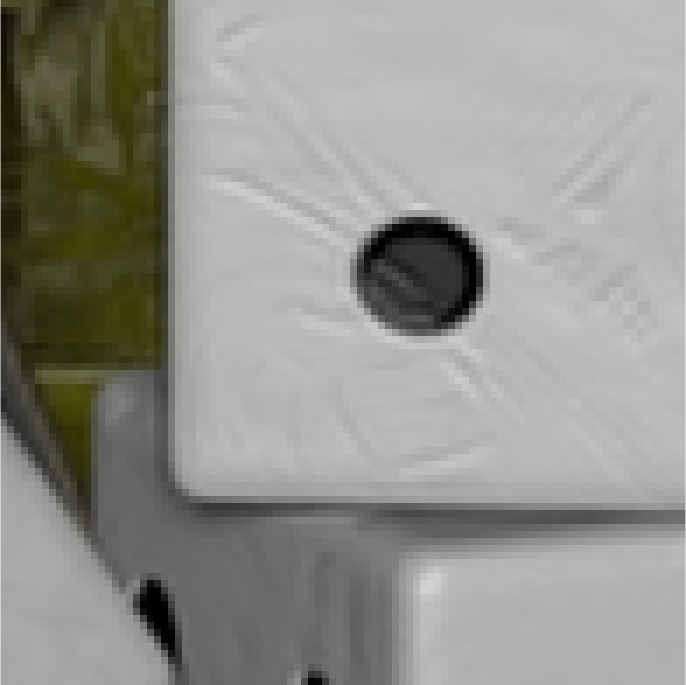}
		\label{fig:hci2_buddhaGBSQ}
	}
	~
	\subfloat[Original HR]{%
		\includegraphics[height=3cm]{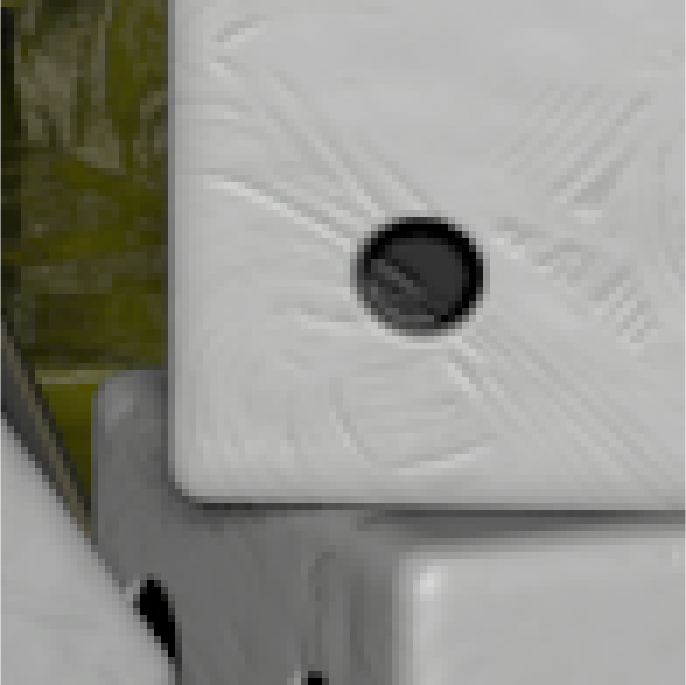}
		\label{fig:hci2_buddhaHR}
	}
	\caption{Detail from the bottom right-most view of the light field \texttt{buddha}, in the HCI dataset.
	The low resolution light field in (a) is super-resolved by a factor ${\alpha = 2}$ with bilinear interpolation in (b),
	the method \cite{wanner_spatial_2012} in (c), the method \cite{mitra_light_2012} in (d),
	the method \cite{dong_learning_2014} in (e), GB-DR in (f) and GB-SQ in (g).
	The original high resolution light field is provided in (h).}
	\label{fig:hci2_buddha}
\end{figure*}
\begin{figure*}
	\centering
	\subfloat[LR]{%
		\includegraphics[height=3cm]{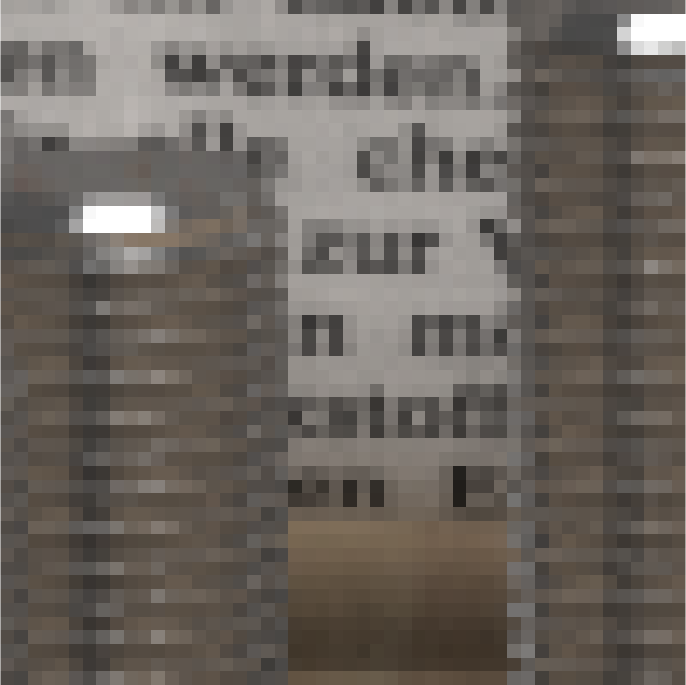}
	}
	~
	\subfloat[Bilinear]{%
		\includegraphics[height=3cm]{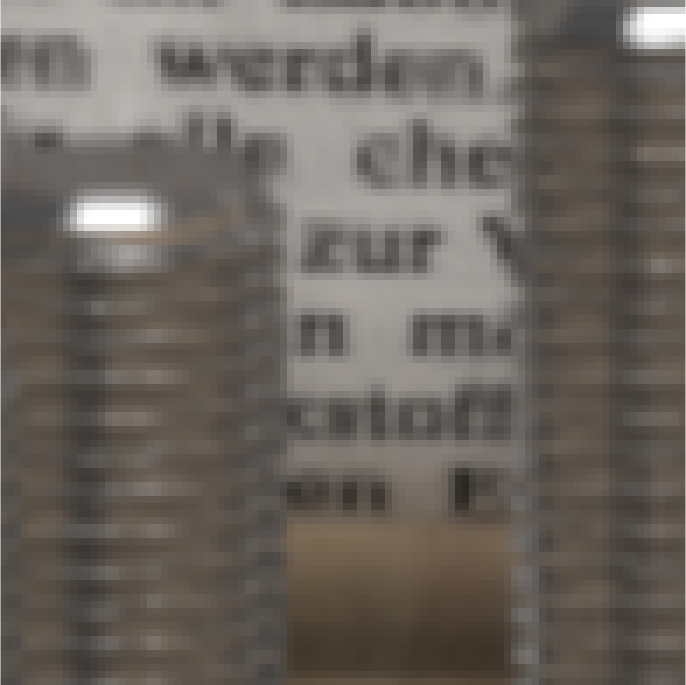}
		\label{fig:hci2_horsesBIL}
	}
	~
	\subfloat[ \cite{wanner_spatial_2012} ]{%
		\includegraphics[height=3cm]{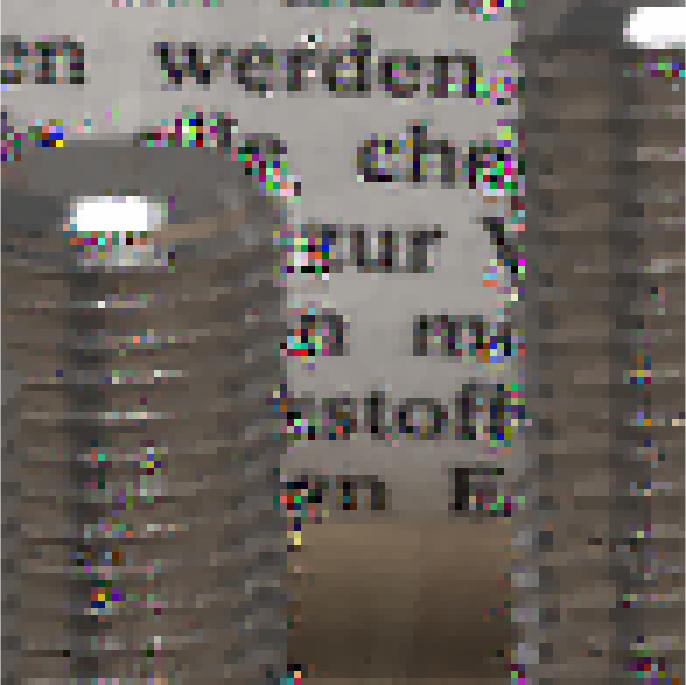}
		\label{fig:hci2_horsesCOCO}
	}
	~
	\subfloat[ \cite{mitra_light_2012} ]{%
		\includegraphics[height=3cm]{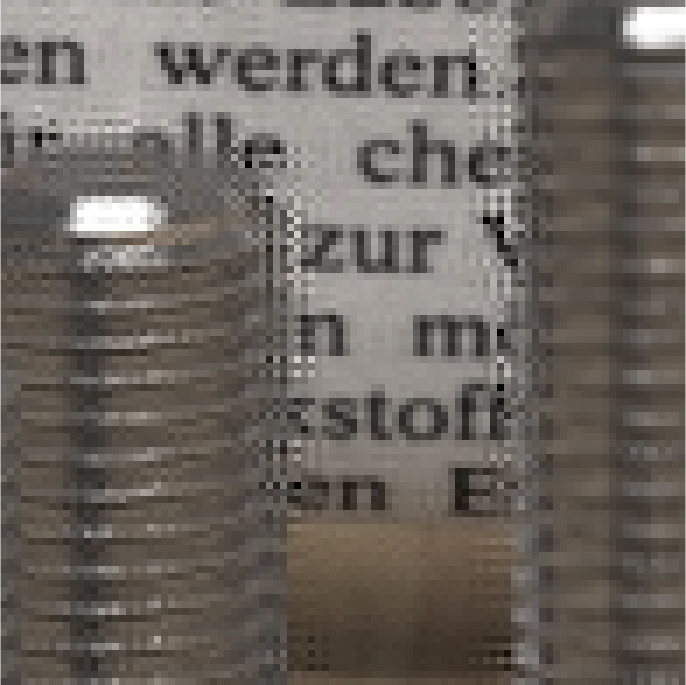}
		\label{fig:hci2_horsesGMM}
	}
	\vspace{0.1cm}
	\subfloat[ \cite{dong_learning_2014} ]{%
		\includegraphics[height=3cm]{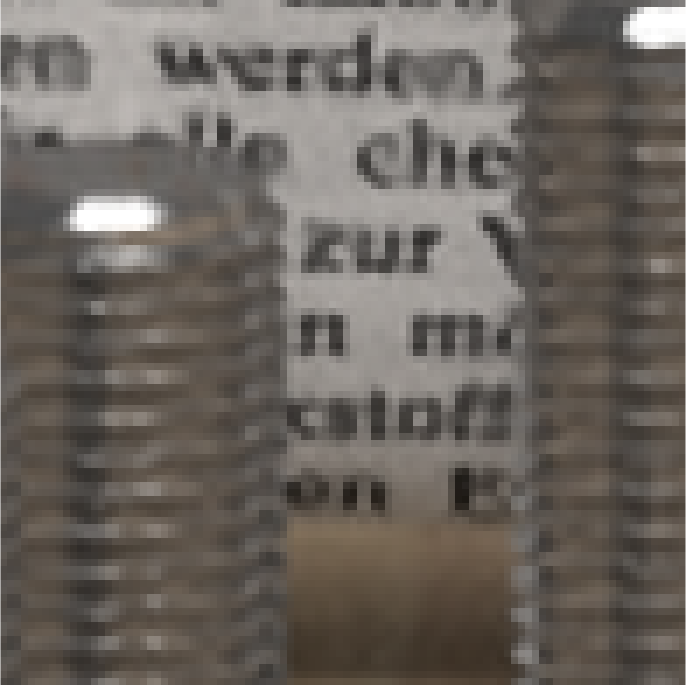}
		\label{fig:hci2_horsesSRCNN}
	}
	~
	\subfloat[GB-DR]{%
		\includegraphics[height=3cm]{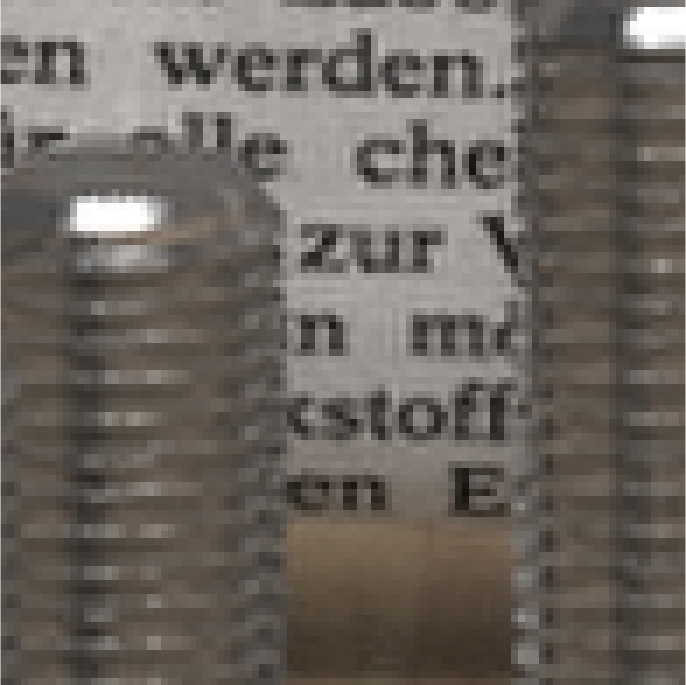}
		\label{fig:hci2_horsesGBDR}
	}
	~
	\subfloat[GB-SQ]{%
		\includegraphics[height=3cm]{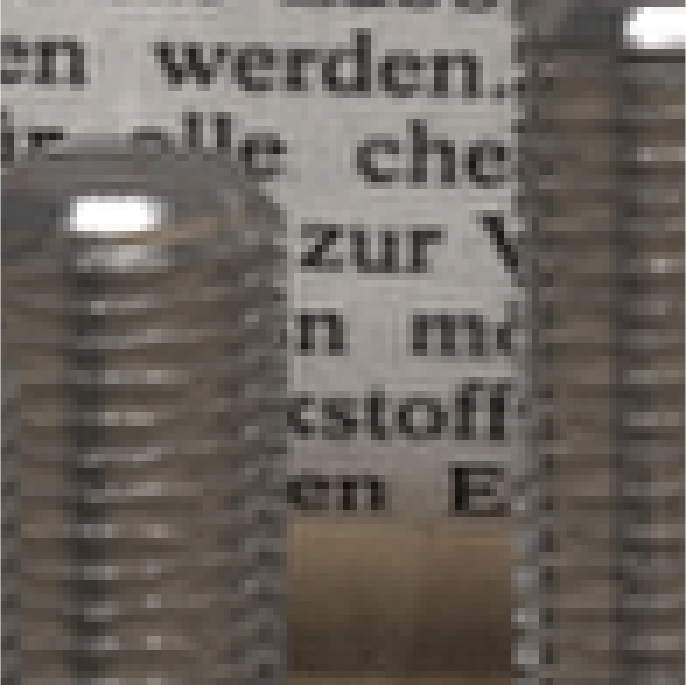}
		\label{fig:hci2_horsesGBSQ}
	}
	~
	\subfloat[Original HR]{%
		\includegraphics[height=3cm]{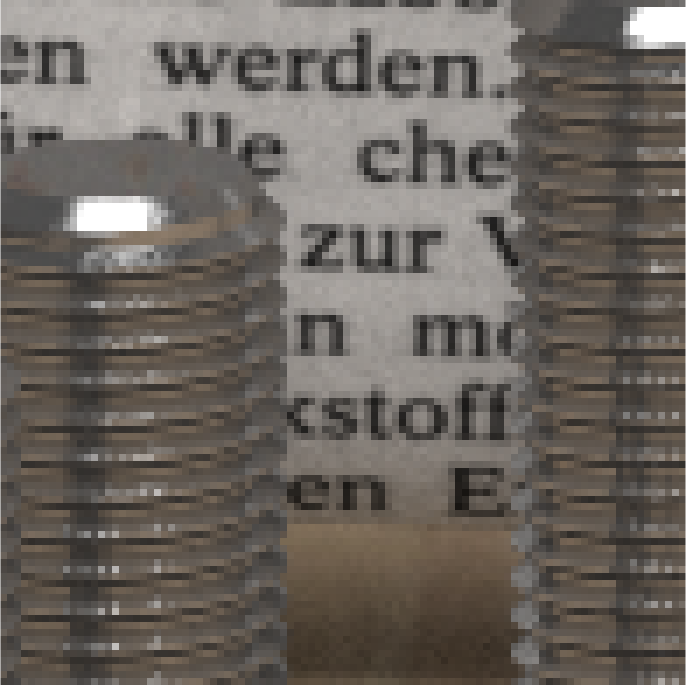}
		\label{fig:hci2_horsesHR}
	}
	\caption{Detail from the bottom right-most view of the light field \texttt{horses}, in the HCI dataset.
	The low resolution light field in (a) is super-resolved by a factor ${\alpha = 2}$ with bilinear interpolation in (b),
	the method \cite{wanner_spatial_2012} in (c), the method \cite{mitra_light_2012} in (d),
	the method \cite{dong_learning_2014} in (e), GB-DR in (f) and GB-SQ in (g).
	The original high resolution light field is provided in (h).}
	\label{fig:hci2_horses}
\end{figure*}

We carry out our experiments on the light field super-resolution algorithms in \cite{wanner_spatial_2012} using the original code by
the authors, available online within the last release of the image processing library \textit{cocolib}.
For a fair comparison, we provide the $[-6, 6]$ pixel range at the library input, in order to permit the removal of outliers in the estimated
disparity maps.
For our experiments on the algorithm in \cite{mitra_light_2012} we use the code provided by the authors.
We discretize the ${[-6, 6]}$ pixel range using a $0.2$ pixel step, and for each disparity value we train a different GMM prior.
The procedure is carried out for ${\alpha = 2}$ and $3$, and results in GMM priors defined on a ${4 \alpha \times 4 \alpha \times M \times M}$
light field patch.
A light field patch is equivalent to a sub-light-field, but with a very small spatial resolution.
We perform the training on the data that comes together with the authors' code.
We also compare GB with a state-of-the-art super-resolution method for single-frame super-resolution, and relying on a CNN \cite{dong_learning_2014}.
For the comparison with the CNN-based super-resolution algorithm in \cite{dong_learning_2014},
we employ the original code from the authors, available online.
We perform the CNN training on the data provided together with the code. In particular, when generating the
\textit{(low, high)-resolution} patch pairs of the training set, we employ the blur and sampling matrix $\bm{SB}$ of Eq.~(\ref{eq:hr2lr})
and previously defined.
We learn one CNN for ${\alpha = 2}$ and one for $3$, and perform ${8 * 10 ^{8}}$ back-propagations, as described in \cite{dong_learning_2014}.
We then super-resolve each light field by applying the trained CNN on the single low resolution views.
Finally, as a baseline reconstruction, we consider also the high resolution light field obtained from bilinear interpolation of the single
low resolution views.

In the experiments, our super-resolution method GB, the methods in \cite{mitra_light_2012} and \cite{dong_learning_2014}, and the
bilinear interpolation one, super-resolve only the luminance of the low resolution light field.
The full color high resolution light field is obtained through bilinear interpolation of the two low resolution light field chrominances.
Instead, for the method in \cite{wanner_spatial_2012}, the corresponding \textit{cocolib} library needs to be fed with the full color low
resolution light field and a full color high resolution light field is provided at the output.
Since most of the considered super-resolution methods super-resolve only the luminance of the low resolution light field,
we compute the reconstruction error only on the luminance channel.
For the method in \cite{wanner_spatial_2012}, whose light field luminance is not available directly, we compute it from the corresponding
full color high resolution light field at the \textit{cocolib} library output.

\subsection{Light fields reconstruction results} \label{subsec:reconstruction_results}

The numerical results from our super-resolution experiments on the HCI and Stanford datasets are reported in Tables~\ref{tab:hci2} and
\ref{tab:stanford2} for a super-resolution factor $\alpha = 2$ and respectively for $\alpha = 3$ in Tables~\ref{tab:hci3} and \ref{tab:stanford3}.
For each reconstructed light field we compute the PSNR (dB) at each view and report the average and variance of the computed PSNRs in the tables.
Finally, for a fair comparison with the method in \cite{mitra_light_2012}, which suffers from border effects, a 15-pixel border is
removed from all the reconstructed views before the PSNR computation.

\begin{figure*}
	\centering
	\begin{tabular}{cc}
	\rotatebox{90}{\makebox[2cm]{Original HR}} &
	\includegraphics[height=2.38cm]{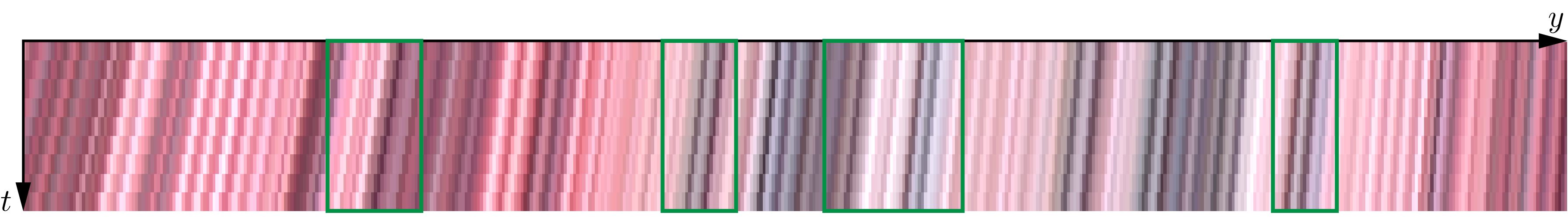} \\
	\rotatebox{90}{\makebox[2cm]{\cite{dong_learning_2014}}} &
	\includegraphics[height=2.38cm]{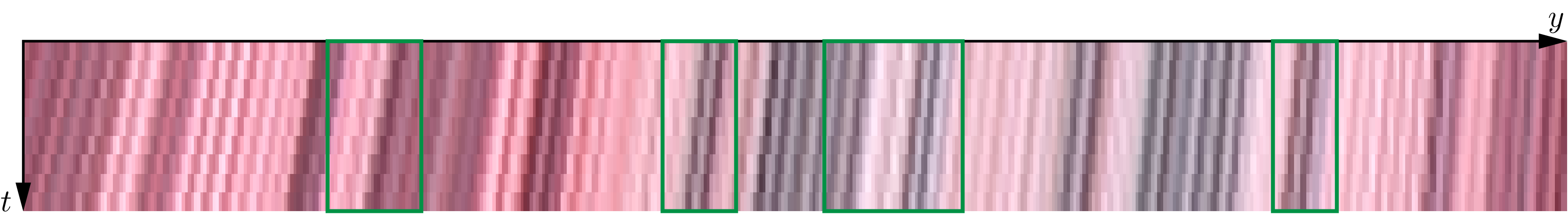} \\
	\rotatebox{90}{\makebox[2cm]{GB-SQ}} &
	\includegraphics[height=2.38cm]{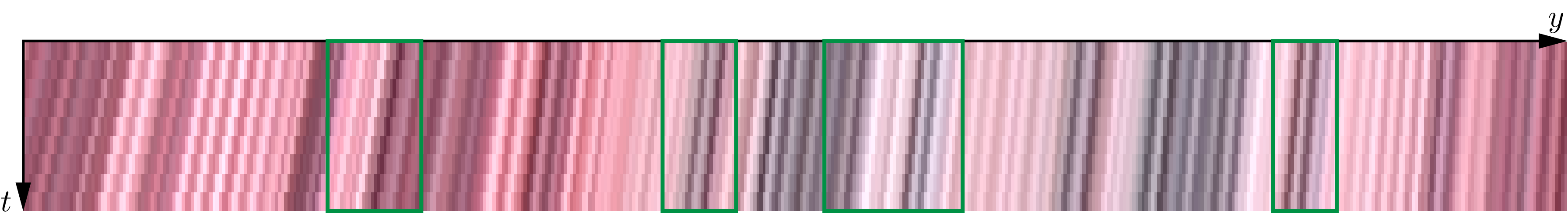}
	\end{tabular}
	\caption{Epipolar image (EPI) from the light field \texttt{stillLife}, in the HCI dataset.
	The ${9 \times 9}$ light field is super-resolved by a factor ${\alpha = 2}$ using the single-frame super-resolution method
	in \cite{dong_learning_2014} and GB-SQ.
	The same EPI is extracted from the original HR light field (top row) and from the reconstructions provided by \cite{dong_learning_2014}
	(central row) and GB-SQ (bottom row).
	Since the method in \cite{dong_learning_2014} super-resolves the views independently, the original line pattern appears compromised,
	therefore the light field structure is not preserved.
	On the contrary, GB-SQ preserves the original line pattern, hence the light field structure.}
	\label{fig:epi_comparison}
\end{figure*}
\begin{figure}
	\centering
	\subfloat[GB-DR]{%
		\includegraphics[height=3.5cm]{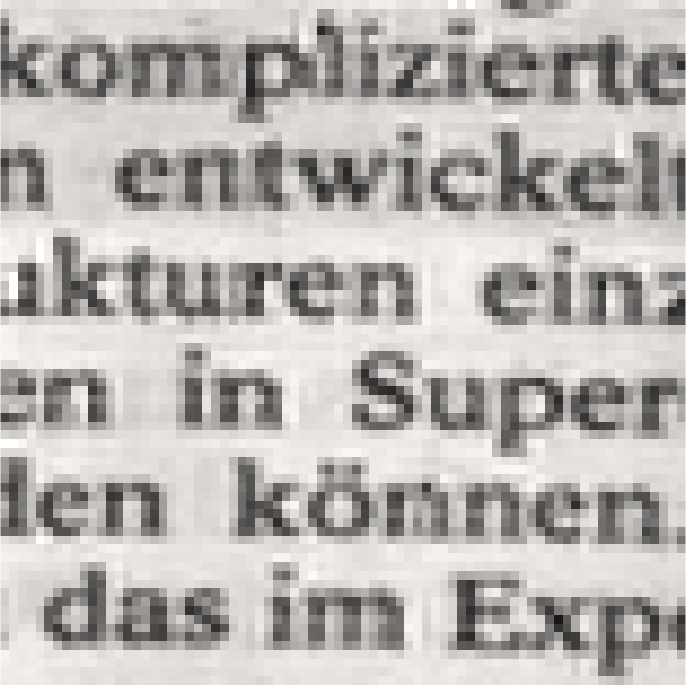}
	}
	~
	\subfloat[GB-SQ]{%
		\includegraphics[height=3.5cm]{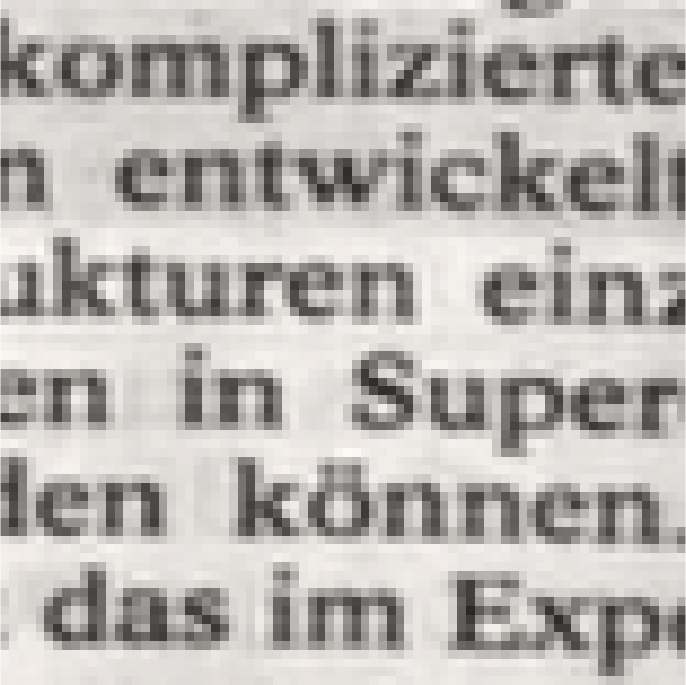}
	}
	\caption{Detail from the central view of the super-resolved light field \texttt{horses}, in the HCI dataset, for the super-resolution factor ${\alpha = 2}$.
	The reconstruction provided by GB-SQ exhibits sharper letters than the reconstruction by GB-DR, as the square constraint captures
	better the light field structure.}
	\label{fig:gb_comparison}
\end{figure}

For a super-resolution factor ${\alpha = 2}$ in the HCI dataset, GB provides the highest average PSNR on ten out of twelve light fields.
In particular, nine out of ten of the highest average PSNRs are due to GB-SQ.
The highest average PSNR in the two remaining light fields \texttt{buddha} and \texttt{horses} is achieved by \cite{mitra_light_2012},
but the corresponding variances are non negligible.
The large variance generally indicates that the quality of the central views is higher than the one of the lateral views.
This is clearly non ideal, as our objective is to reconstruct all the views with high quality, as necessary in most light field applications.
We also note that GB provides a better visual quality in these two light fields.
This is shown in Figure~\ref{fig:hci2_buddha} and \ref{fig:hci2_horses}, where two details from the bottom right-most views of the
light fields \texttt{buddha} and \texttt{horses}, respectively, are given for each method.
In particular, the reconstruction provided by \cite{mitra_light_2012} exhibits strong artifacts along object boundaries.
This method assumes a constant disparity within each light field patch that it processes, but patches capturing object boundaries are
characterized by an abrupt change of disparity that violates this assumption and causes unpleasant artifacts.
Figures~\ref{fig:hci2_buddhaCOCO} and \ref{fig:hci2_horsesCOCO} show that also the reconstructions provided by the method
in \cite{wanner_spatial_2012} exhibit strong artifacts along edges, although the disparity is estimated at each pixel in this case.
This is due to the presence of errors in the estimated disparity at object boundaries.
These errors are caused both by the poor performance of the tensor structure operator in the presence of occlusions,
and more in general to the challenges posed by disparity estimation at low resolution.
We also observe that the TV term in \cite{wanner_spatial_2012} tends to over-smooth the fine details, as evident in the dice of
Figure~\ref{fig:hci2_buddhaCOCO}.
The method in \cite{dong_learning_2014}, meant for single-frame super-resolution and therefore agnostic of the light field structure,
provides PSNR values that are significantly lower than those provided by GB and \cite{mitra_light_2012}, which instead take the light field
structure into account.
In particular, the quality of the views reconstructed by the method in \cite{dong_learning_2014} depends exclusively on the training data,
as it does not employ the complementary information available at the other views.
This is clear in Figure~\ref{fig:hci2_buddhaSRCNN}, where \cite{dong_learning_2014} does not manage to recover the fine structure around the
black spot in the dice, which remains pixelated as in the original low resolution view.
Similarly, the method in \cite{dong_learning_2014} does not manage to reconstruct effectively the letters in Figure~\ref{fig:hci2_horsesSRCNN},
which remain blurred and in some cases cannot be discerned.
Moreover, since the method in \cite{dong_learning_2014} does not consider the light field structure, it does not necessarily preserve it.
An example is provided in Figure~\ref{fig:epi_comparison}, where an epipolar image is extracted from the reconstructions of the \texttt{stillLife} light field
computed by GB-SQ and the method in \cite{dong_learning_2014}.
While GB-SQ preserves the line patterns, the method in \cite{dong_learning_2014} does not.
The bilinear interpolation method provides the lowest PSNR values and the poor quality of its reconstruction is confirmed by the
Figures~\ref{fig:hci2_buddhaBIL} and \ref{fig:hci2_horsesBIL}, which appear significantly blurred.
In particular, the fine structure around the black spot in the dice of Figure~\ref{fig:hci2_buddhaHR} is almost absent in the reconstruction
provided by the bilinear interpolation method, and some letters in Figure~\ref{fig:hci2_horsesBIL} cannot be discerned.
Finally, the numerical results suggest that our GB-SQ methods is more effective in capturing the correct correspondences between
adjacent views in the light field.
A visual example is provided in Figure~\ref{fig:gb_comparison}, where the letters in the view reconstructed by GB-SQ are sharper
than those in the view reconstructed by GB-DR.

\begin{figure*}
	\centering
	\subfloat[LR]{%
		\begin{tabular*}{0.1 \textwidth}{c}
		\includegraphics[width=0.095 \textwidth]{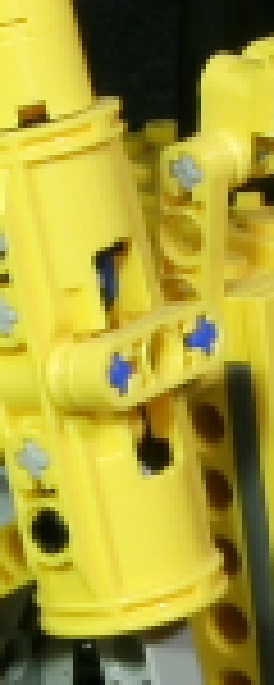} \\
		\includegraphics[width=0.095 \textwidth]{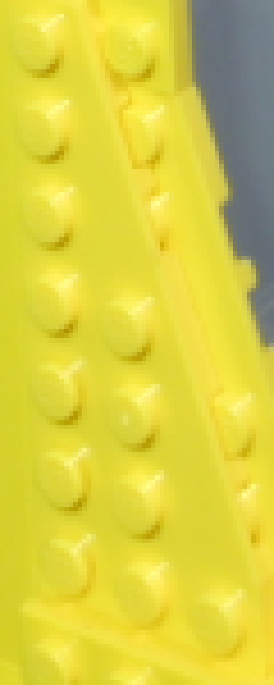}
		\end{tabular*}
	}
	\hfill
	\subfloat[Bilinear]{%
		\begin{tabular*}{0.1 \textwidth}{c}
		\includegraphics[width=0.095 \textwidth]{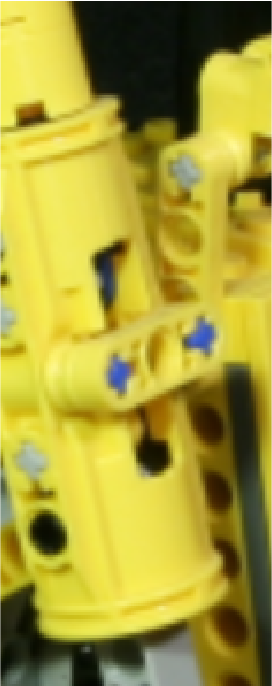} \\
		\includegraphics[width=0.095 \textwidth]{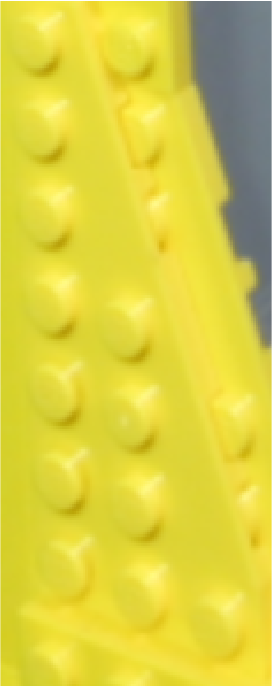}
		\end{tabular*}
	}
	\hfill
	\subfloat[ \cite{wanner_spatial_2012} ]{%
		\begin{tabular*}{0.1 \textwidth}{c}
		\includegraphics[width=0.095 \textwidth]{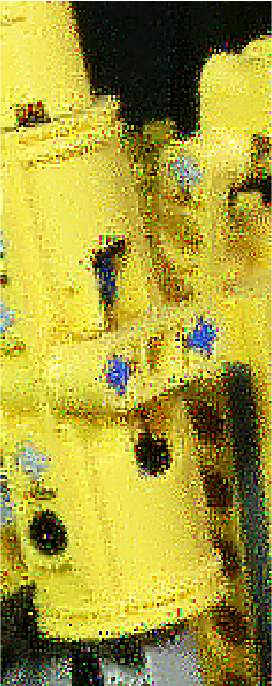} \\
		\includegraphics[width=0.095 \textwidth]{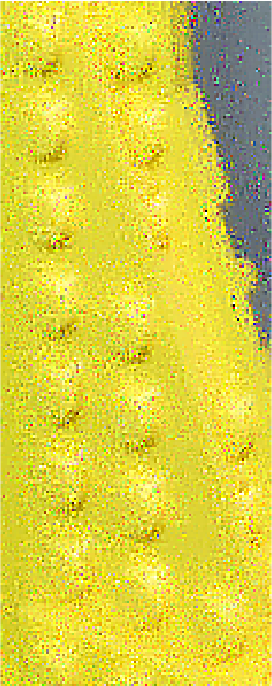}
		\end{tabular*}
	}
	\hfill
	\subfloat[ \cite{mitra_light_2012} ]{%
		\begin{tabular*}{0.1 \textwidth}{c}
		\includegraphics[width=0.095 \textwidth]{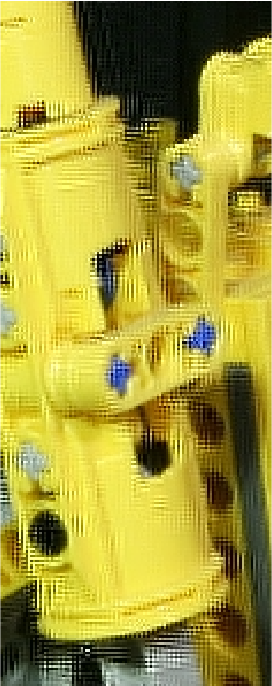} \\
		\includegraphics[width=0.095 \textwidth]{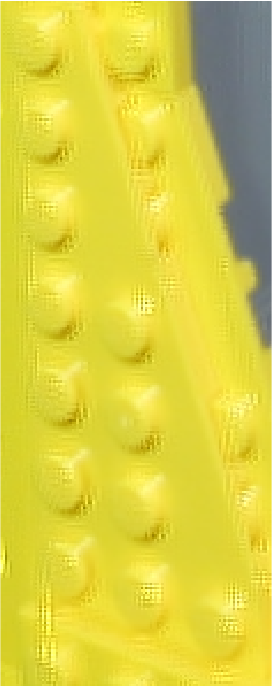}
		\end{tabular*}
		\label{fig:stanford2_bulldozerGMM}
	}
	\hfill
	\subfloat[ \cite{dong_learning_2014} ]{%
		\begin{tabular*}{0.1 \textwidth}{c}
		\includegraphics[width=0.095 \textwidth]{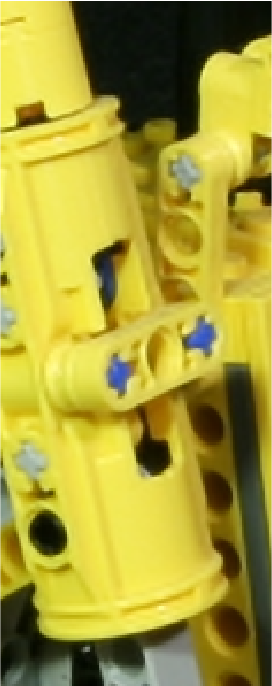} \\
		\includegraphics[width=0.095 \textwidth]{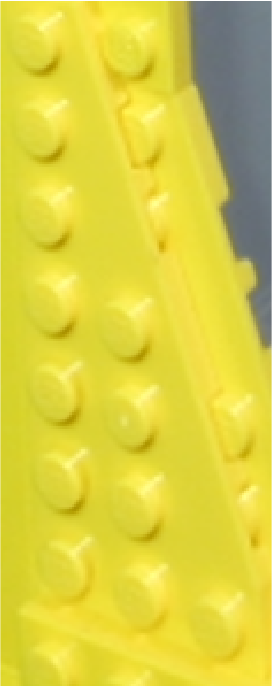}
		\end{tabular*}
	}
	\hfill
	\subfloat[GB-DR]{%
		\begin{tabular*}{0.1 \textwidth}{c}
		\includegraphics[width=0.095 \textwidth]{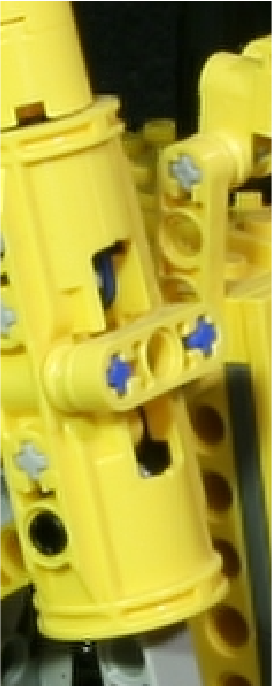} \\
		\includegraphics[width=0.095 \textwidth]{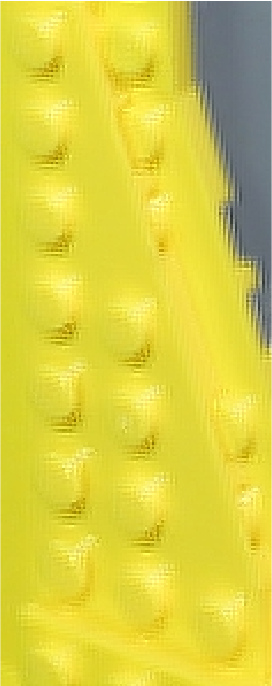}
		\end{tabular*}
	}
	\hfill
	\subfloat[GB-SQ]{%
		\begin{tabular*}{0.1 \textwidth}{c}
		\includegraphics[width=0.095 \textwidth]{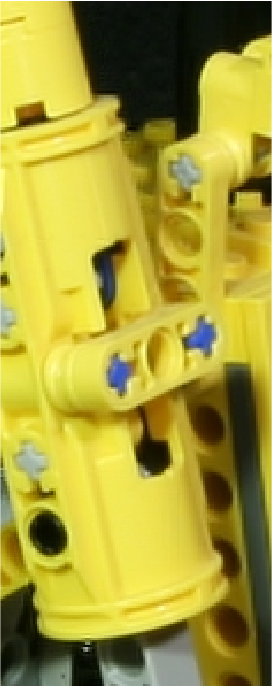} \\
		\includegraphics[width=0.095 \textwidth]{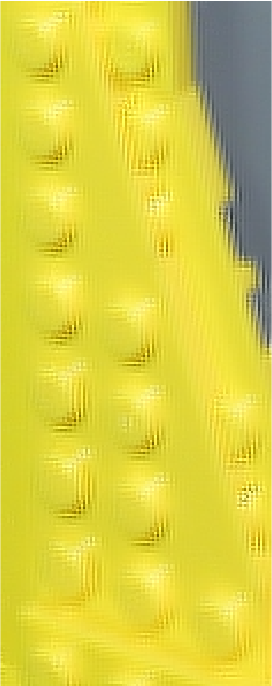}
		\end{tabular*}
	}
	\hfill
	\subfloat[GB-SQ-12]{%
		\begin{tabular*}{0.1 \textwidth}{c}
		\includegraphics[width=0.095 \textwidth]{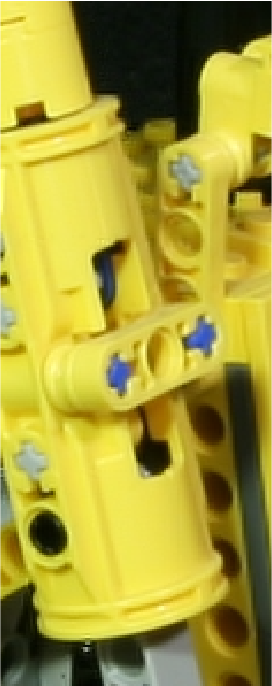} \\
		\includegraphics[width=0.095 \textwidth]{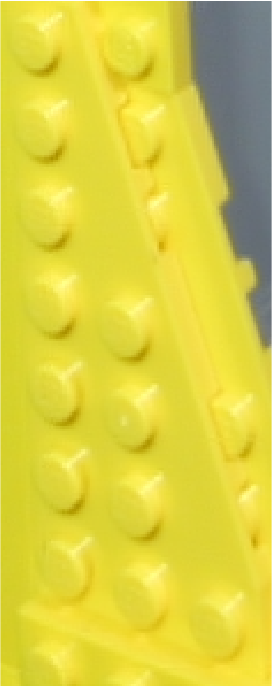}
		\end{tabular*}
		\label{fig:stanford2_bulldozerGBSQ12}
	}
	\hfill
	\subfloat[Original HR]{%
		\begin{tabular*}{0.1 \textwidth}{c}
		\includegraphics[width=0.095 \textwidth]{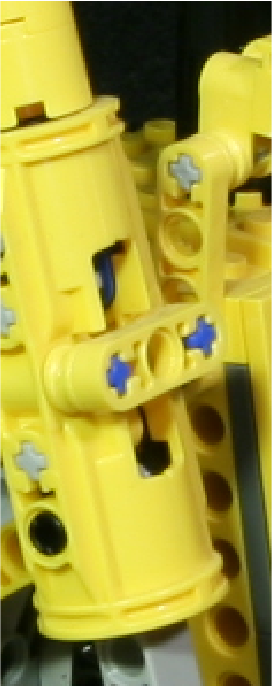} \\
		\includegraphics[width=0.095 \textwidth]{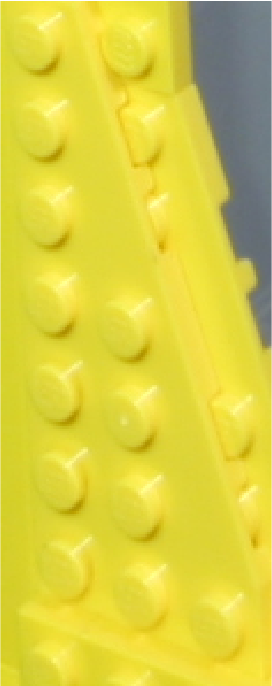}
		\end{tabular*}
	}
	\caption{Details from the bottom right-most view of the light field \texttt{bulldozer}, in the Stanford dataset.
	The low resolution light field in (a) is super-resolved by a factor ${\alpha = 2}$ with bilinear interpolation in (b),
	the method \cite{wanner_spatial_2012} in (c), the method \cite{mitra_light_2012} in (d),
	the method \cite{dong_learning_2014} in (e), GB-DR in (f), and GB-SQ in (g).
	The reconstruction of GB-SQ with the extended disparity range ${[-12,12]}$ pixels is provided in (h),
	and the original high resolution light field is in (i).}
	\label{fig:stanford2_bulldozer}
\end{figure*}

In the Stanford dataset and for the same super-resolution factor ${\alpha = 2}$, GB provides the highest average PSNRs on eight
light fields out of eleven, the method in \cite{dong_learning_2014} provides the highest average PSNRs in the three remaining light fields,
while the algorithms in \cite{wanner_spatial_2012} and \cite{mitra_light_2012} perform even worse than bilinear interpolation in most
of the cases.
The very poor performance of \cite{wanner_spatial_2012} and \cite{mitra_light_2012}, and the generally higher PSNR provided by GB-DR
compared to GB-SQ, are mainly consequences of the Stanford dataset disparity range, which exceeds the ${[-6,6]}$ pixel range assumed
in our tests.
In particular, objects with a disparity outside the assumed disparity range are not properly reconstructed in general.
An example is provided in Figure~\ref{fig:stanford2_bulldozer}, where two details from the bottom right-most view of the
light field \texttt{bulldozer} are shown.
The detail at the bottom captures the bulldozer blade, placed very close to the camera and characterized by large disparity values
outside the assumed disparity range, while the detail on the top captures a cylinder behind the blade and characterized by disparity 
values within the assumed range.
As expected, GB manages to correctly reconstruct the cylinder, while it introduces some artifacts on the blade.
However, it can be observed that GB-DR introduces milder artifacts than GB-SQ on the blade, as GB-SQ forces the warping matrices to fulfill
the square constraint of Section~\ref{sec:warp_matrices} on a wrong disparity range, while GB-DR is more accommodating
in the warping matrix construction and therefore more robust to a wrong disparity range assumption.
For the sake of completeness, Figure~\ref{fig:stanford2_bulldozerGBSQ12} provides the reconstruction computed by GB-SQ when the
assumed disparity range is extended to ${[-12,12]}$ pixels, and it shows that the artifacts disappear when the correct disparity range
is within the assumed one.
On the other hand, in Figure~\ref{fig:stanford2_bulldozerGMM} the method in \cite{mitra_light_2012} fails to reconstruct also the cylinder,
as the top of the image exhibits depth discontinuities that do not fit its assumption of constant disparity within each light field patch.
The method in \cite{wanner_spatial_2012} fails in both areas as well, and in general on the whole Stanford light field dataset,
as the structure tensor operator cannot detect large disparity values \cite{diebold_refocus}.
Differently from the light-field super-resolution methods, the one in \cite{dong_learning_2014} processes each view independently
and it does not introduce any visible artifact, neither in the top nor in the bottom detail.
However, the absence of visible artifacts does not guarantee that the light field structure is preserved, as \cite{dong_learning_2014}
does not take it into account.
For the sake of completeness, we observe that not all the light fields in the Stanford dataset meet the Lambertian assumption.
Some areas of the captured scenes violate it.
This contributes to the low PSNR values exhibited by the methods \cite{wanner_spatial_2012}, \cite{mitra_light_2012}, and GB-SQ,
on certain light fields (e.g., \texttt{bracelet}) in Table~\ref{tab:stanford2}, as in non Lambertian areas the light field structure
in Eq.~(\ref{eq:twodim_stereo}) does not hold true.
On the other hand, as we already stated, GB-DR is more accommodating in the warping matrix construction and this makes the method
more robust not only to the adoption of incorrect disparity ranges, but also to the violation of the Lambertian assumption, as confirmed
numerically in Table~\ref{tab:stanford2}.

\begin{figure*}
	\centering
	\subfloat[LR]{%
		\includegraphics[height=3cm]{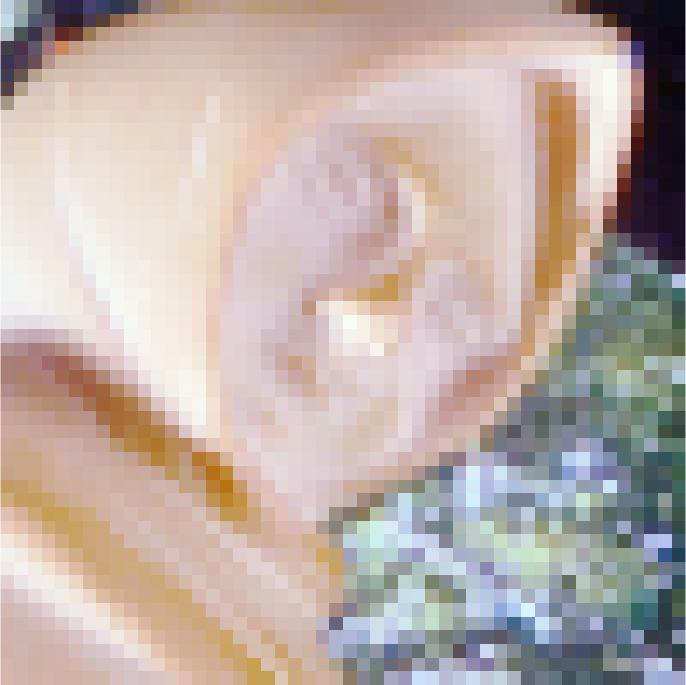}
	}
	~
	\subfloat[Bilinear]{%
		\includegraphics[height=3cm]{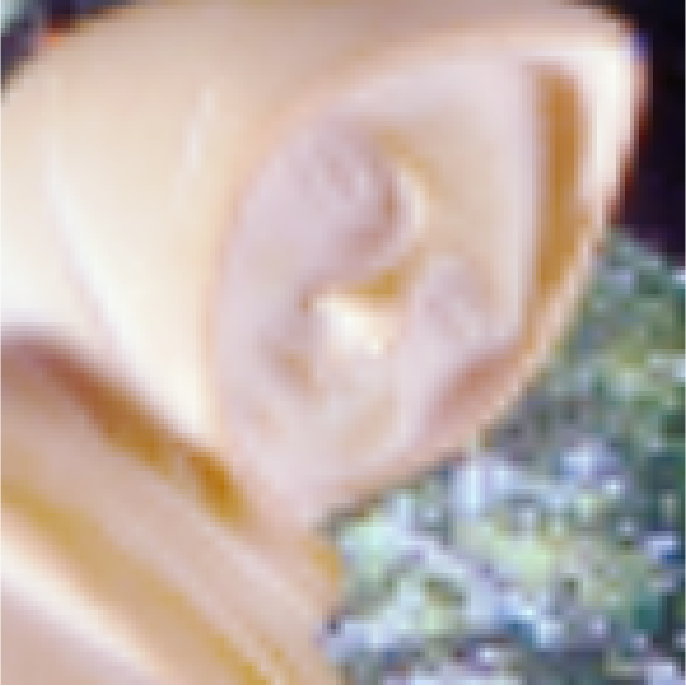}
		\label{fig:hci3_statueBIL}
	}
	~
	\subfloat[ \cite{wanner_spatial_2012} ]{%
		\includegraphics[height=3cm]{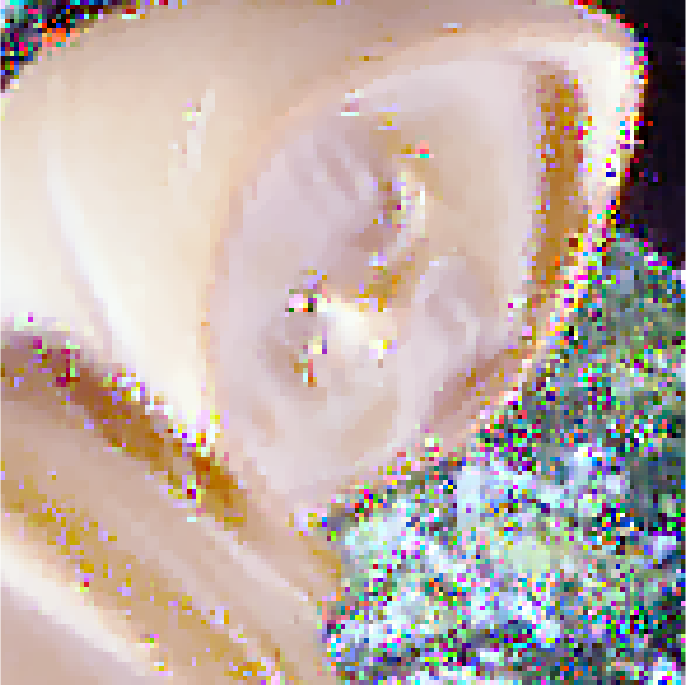}
	}
	~
	\subfloat[ \cite{mitra_light_2012} ]{%
		\includegraphics[height=3cm]{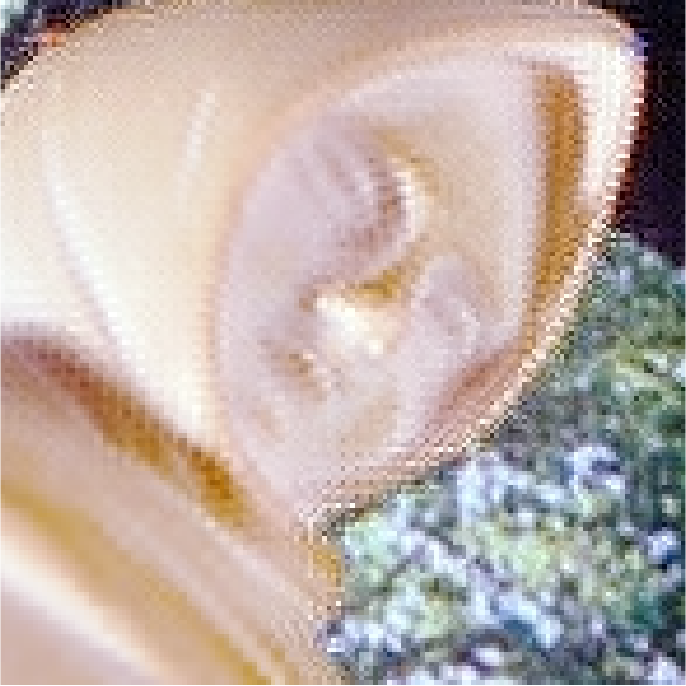}
	}
	\vspace{0.1cm}
	\subfloat[ \cite{dong_learning_2014} ]{%
		\includegraphics[height=3cm]{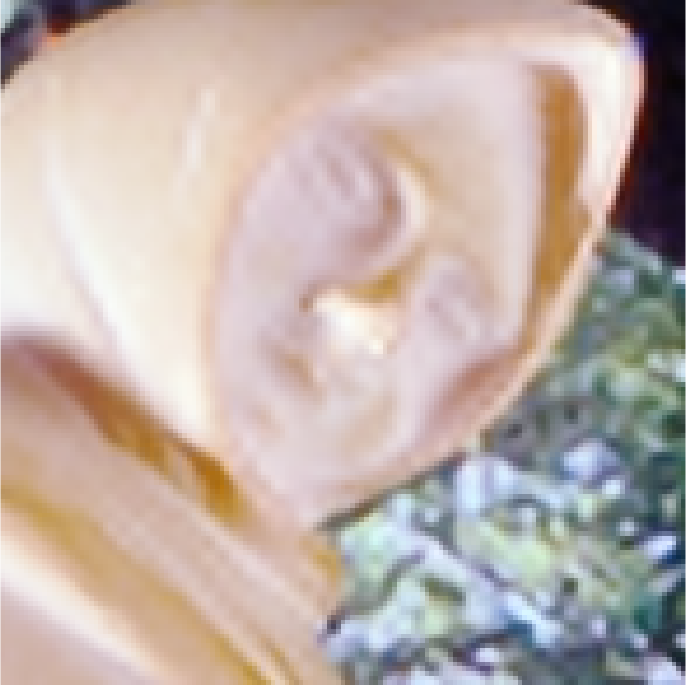}
	}
	~
	\subfloat[GB-DR]{%
		\includegraphics[height=3cm]{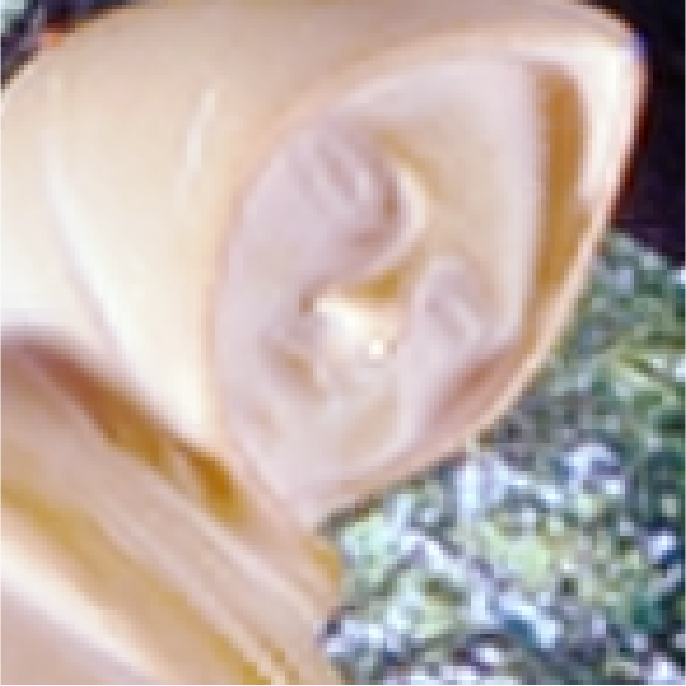}
	}
	~
	\subfloat[GB-SQ]{%
		\includegraphics[height=3cm]{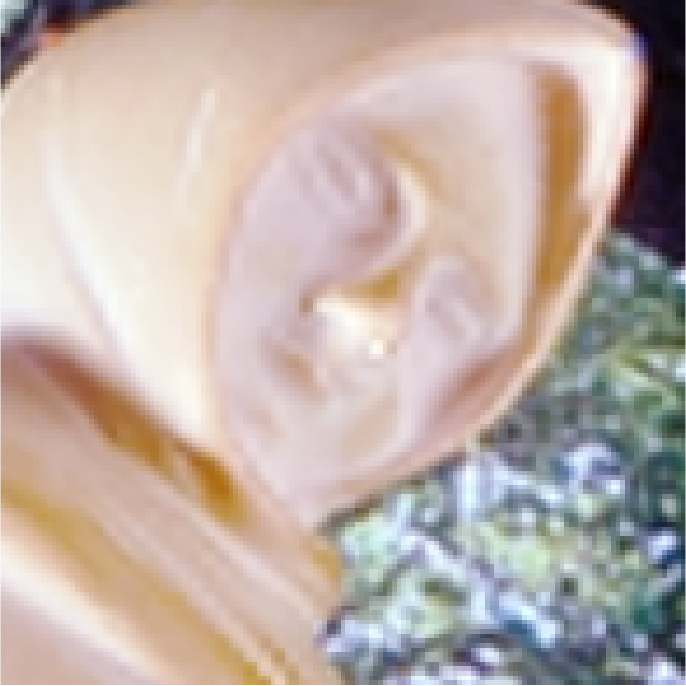}
	}
	~
	\subfloat[Original HR]{%
		\includegraphics[height=3cm]{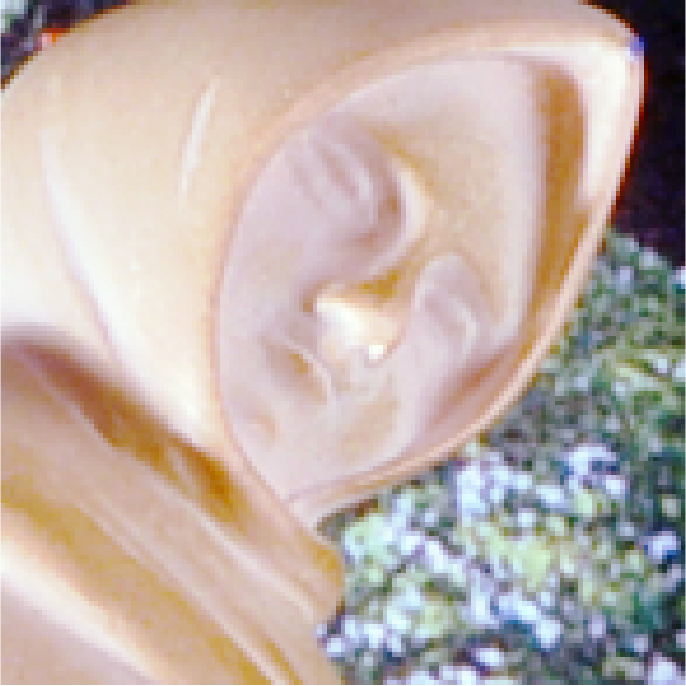}
	}
	\caption{Detail from the bottom right-most view of the light field \texttt{statue}, in the HCI dataset.
	The low resolution light field in (a) is super-resolved by a factor ${\alpha = 3}$ with bilinear interpolation in (b),
	the method \cite{wanner_spatial_2012} in (c), the method \cite{mitra_light_2012} in (d),
	the method \cite{dong_learning_2014} in (e), GB-DR in (f) and GB-SQ in (g).
	The original high resolution light field is provided in (h).}
	\label{fig:hci3_statue}
\end{figure*}

We now consider a larger super-resolution factor of ${\alpha = 3}$.
In the HCI dataset, the method in \cite{mitra_light_2012} provides the highest average PSNRs on half of the light fields,
while GB provides the highest average PSNRs only on four of them.
However, the average PSNR happens to be a very misleading index here.
In particular, the method in \cite{mitra_light_2012} provides the highest average PSNR on the light field \texttt{statue},
but the PSNR variance is larger than $2$~dB, which indicates a very large difference in the quality of the reconstructed images.
On the other hand, GB-SQ provides a slightly lower average PSNR on the same light field, but the PSNR variance is $0.01$~dB,
which suggests a more homogenous quality of the reconstructed light field views.
In particular, the lowest PNSR provided by GB-SQ among all the views is equal to $28.21$~dB, which is almost 3 dB higher than
the worst case view reconstructed by \cite{mitra_light_2012}.
Moreover, the light fields reconstructed by \cite{mitra_light_2012} exhibit very strong artifacts along object boundaries.
An example is provided in Figure~\ref{fig:hci3_statue}, which represents a detail from the central view of the light field \texttt{statue}.
The head of the statue reconstructed by \cite{mitra_light_2012} appears very noisy, especially at the depth discontinuity between the
head and the background, while GB is not significantly affected.
The lower average PSNR provided by GB on some light field, when compared to \cite{mitra_light_2012}, is caused by the very poor
resolution of the input data for $\alpha = 3$, that makes the capture of the correct matches for the warping matrix construction more
and more challenging.
However, as suggested by Figure~\ref{fig:hci3_statue}, the regularizer $\mathcal{F} _{3}$ manages to compensate for these errors.
The method in \cite{wanner_spatial_2012} performs worse than \cite{mitra_light_2012} and GB both in terms of PSNR and visually.
As an example, in Figure~\ref{fig:hci3_statue} the reconstruction provided by \cite{wanner_spatial_2012} shows strong artifacts not
only at depth discontinuities, but especially in the background, which consists of a flat panel with a tree motive.
Despite the very textured background, the tensor structure fails to capture the correct depth due to the very low resolution of the views,
and this has a dramatic impact on the final reconstruction.
In general, depth estimation at very low resolution happens to be a very challenging task.
The method in \cite{dong_learning_2014} reconstructs the statue of Figure~\ref{fig:hci3_statue} correctly, and no unpleasant artifacts are visible.
However, it introduces new structures in the textured background and this leads the PSNR to drop.
In general, the method in \cite{dong_learning_2014}  provides lower average PSNR values than GB on the twelve light fields, as the separate
processing of each views makes it agnostic of the complementary information in the others and it can rely only on the data it scanned
in the training phase.
Finally, the worst numerical results are provided mainly by the bilinear interpolation method, which does not exhibit strong artifacts in general,
but provides very blurred images, as shown in Figure~\ref{fig:hci3_statueBIL} and expected.

Finally, for the Stanford dataset and ${\alpha = 3}$, the numerical results in Table~\ref{tab:stanford3} show a similar behavior to the one
observed for ${\alpha = 2}$.
The methods in \cite{wanner_spatial_2012} and \cite{mitra_light_2012} are heavily affected by artifacts, due to the disparities exceeding
the assumed range.
Instead, GB proves to be more robust to the incorrect disparity range, in particular the variant GB-DR.
The method in \cite{dong_learning_2014} is limited by its considering the views separately, although it is not affected by the artifacts
caused by the incorrect disparity range.

\begin{figure*}
	\centering
	\begin{tabular}{cccccc}
	\rotatebox{90}{\makebox[3cm]{Original LR}} &
	\includegraphics[height=3cm]{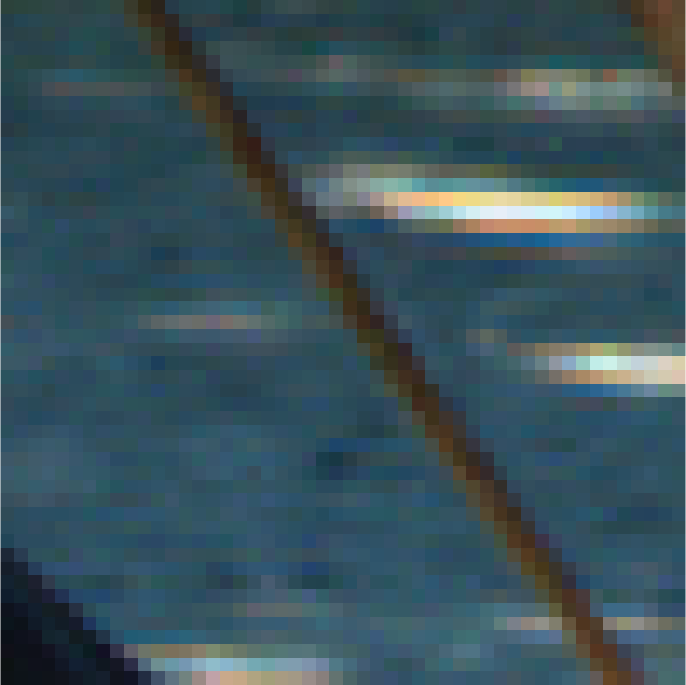} &
	\includegraphics[height=3cm]{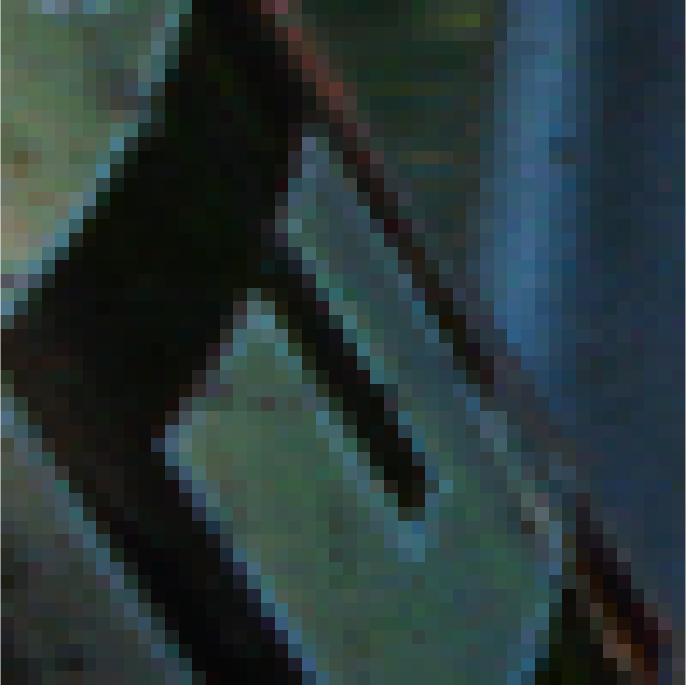} &
	\includegraphics[height=3cm]{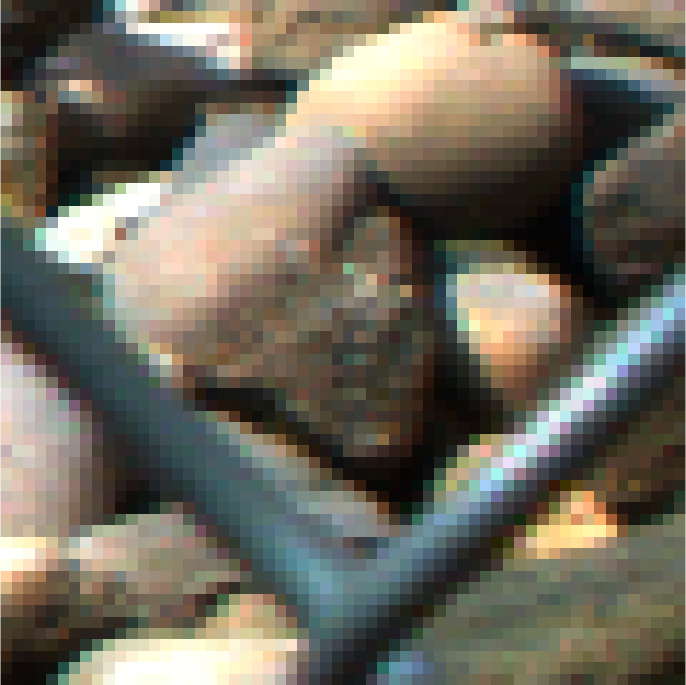} &
	\includegraphics[height=3cm]{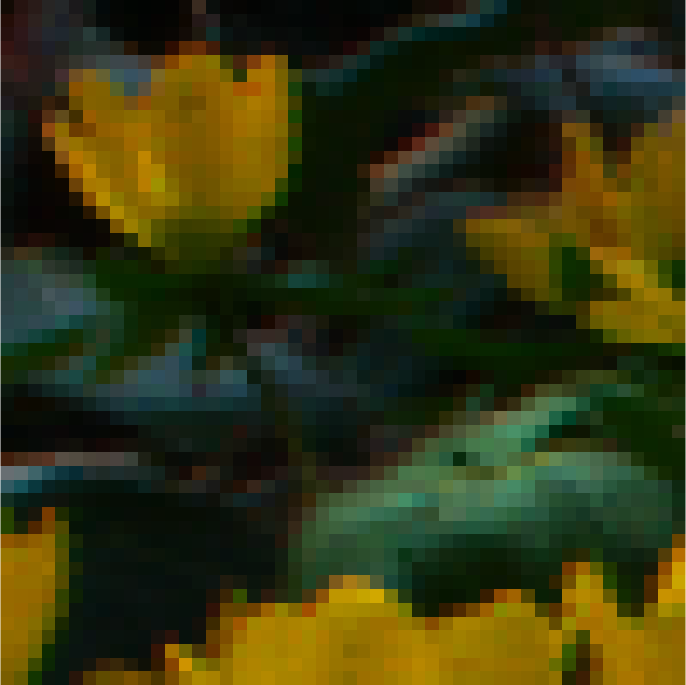} &
	\includegraphics[height=3cm]{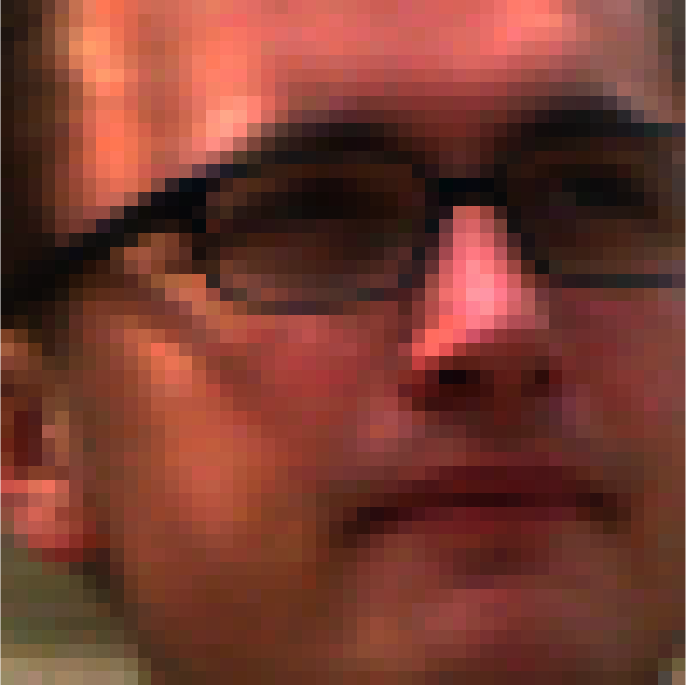} \\
	\rotatebox{90}{\makebox[3cm]{\cite{mitra_light_2012}}} &
	\includegraphics[height=3cm]{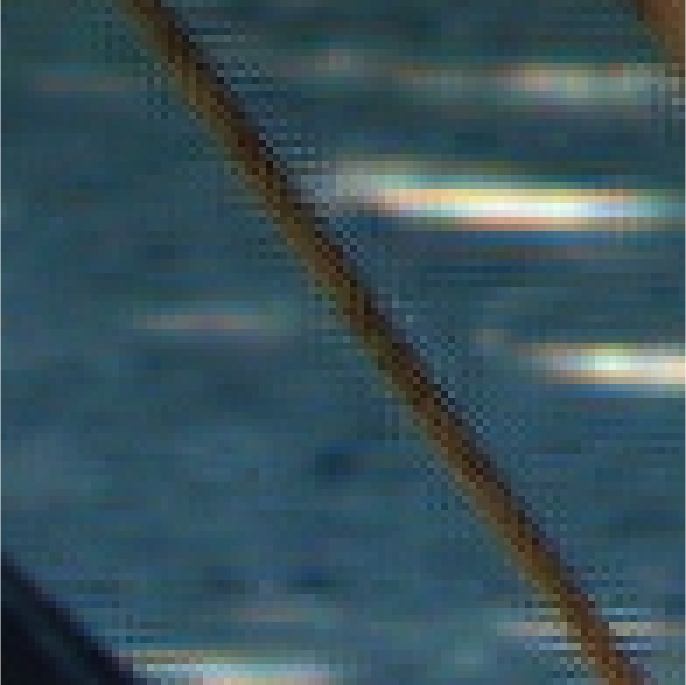} &
	\includegraphics[height=3cm]{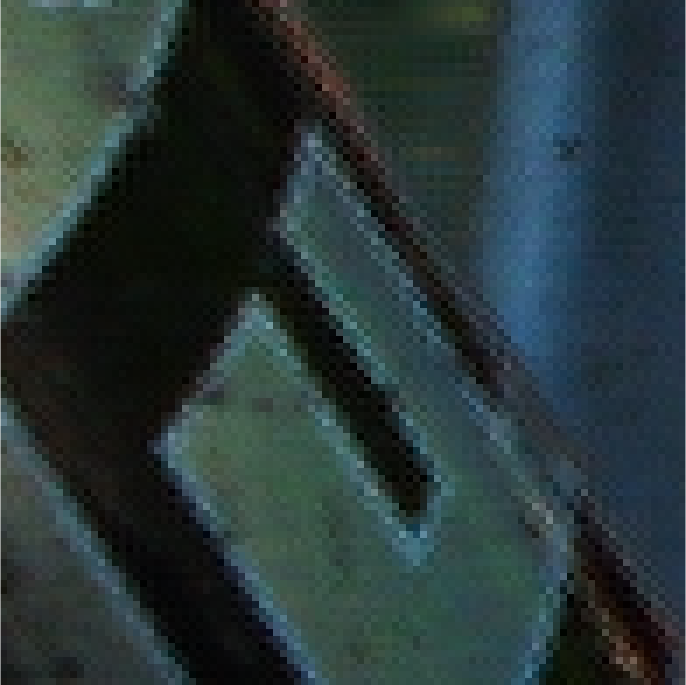} &
	\includegraphics[height=3cm]{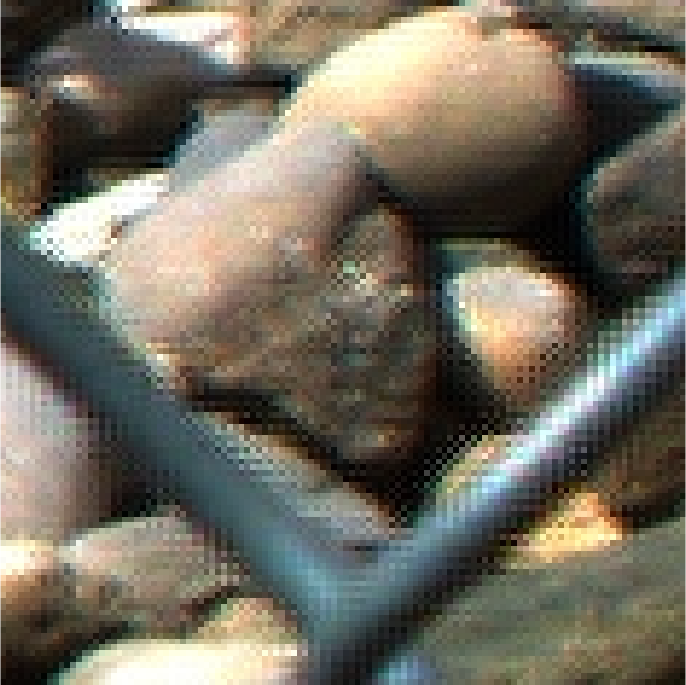} &
	\includegraphics[height=3cm]{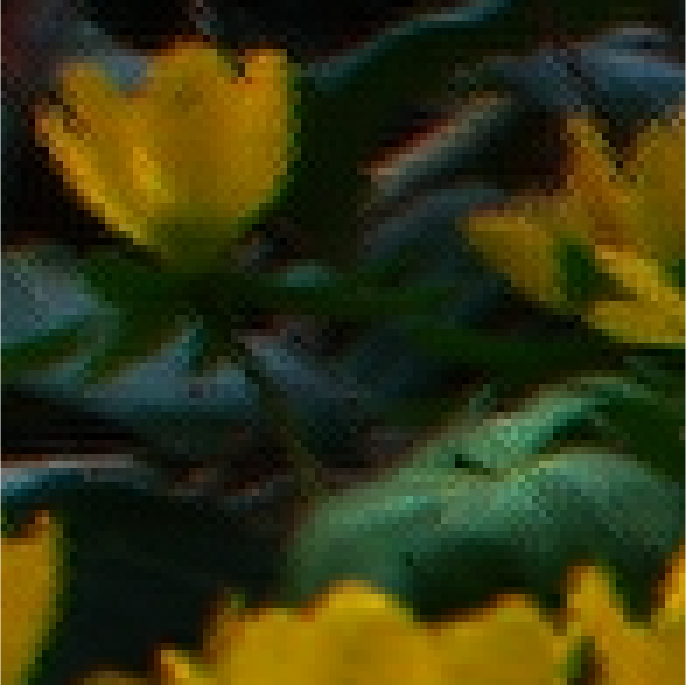} &
	\includegraphics[height=3cm]{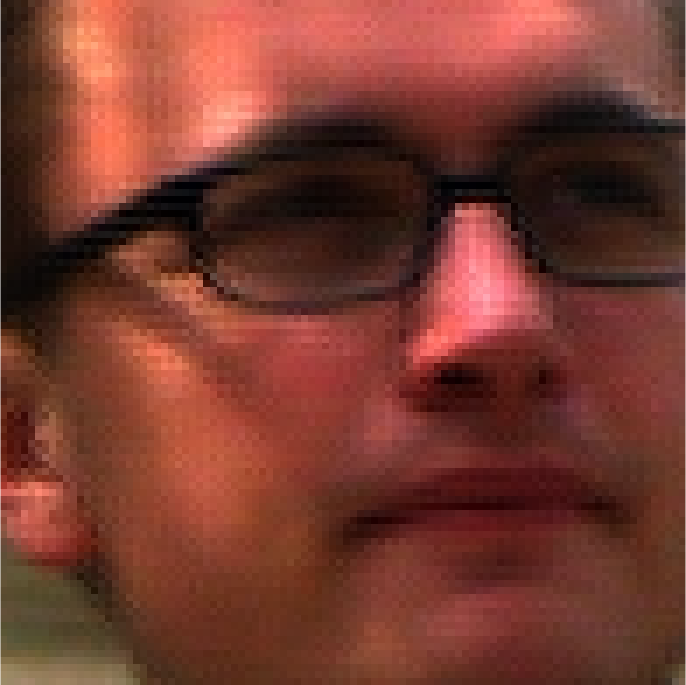} \\
	\rotatebox{90}{\makebox[3cm]{GB-SQ}} &
	\includegraphics[height=3cm]{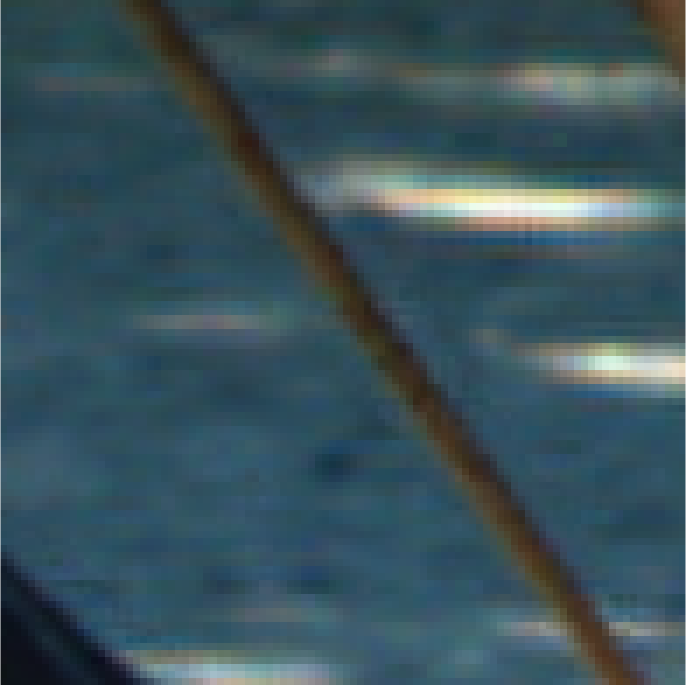} &
	\includegraphics[height=3cm]{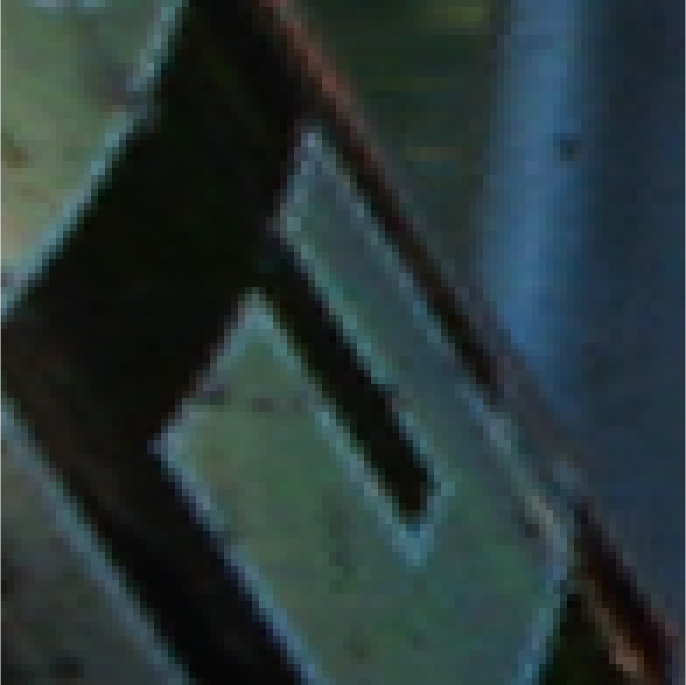} &
	\includegraphics[height=3cm]{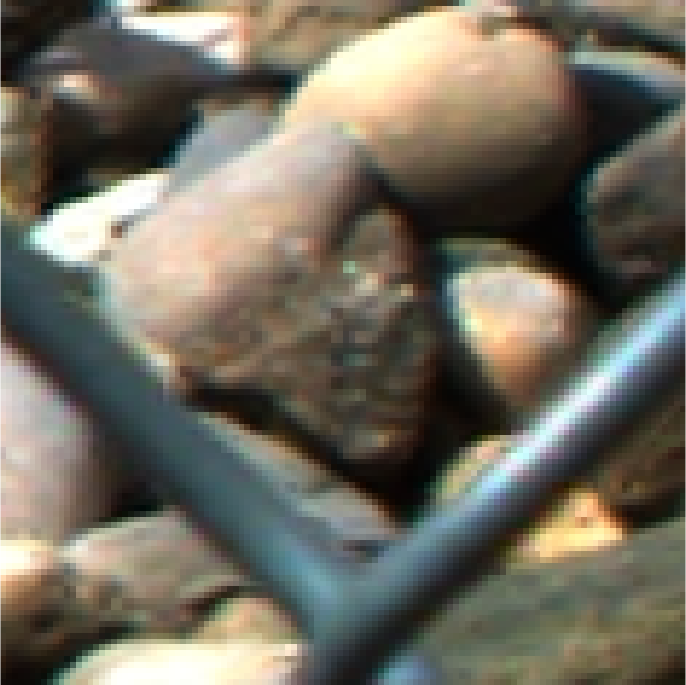} &
	\includegraphics[height=3cm]{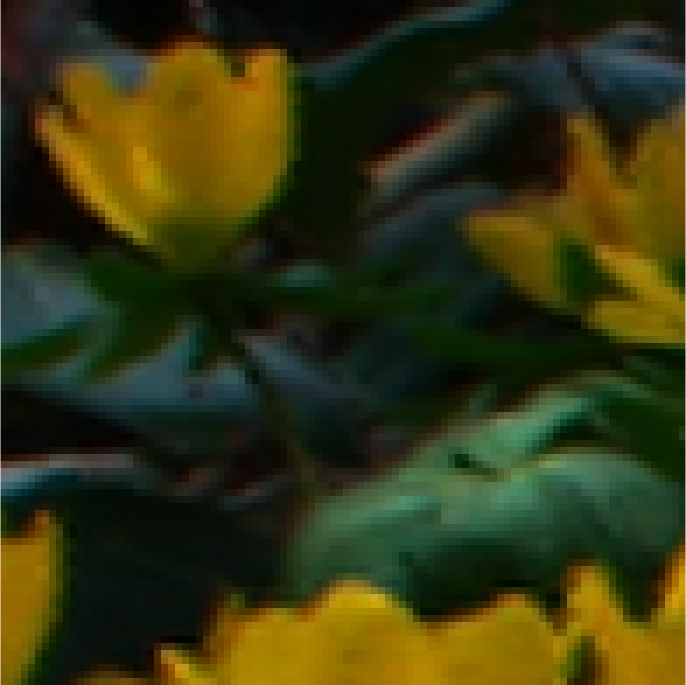} &
	\includegraphics[height=3cm]{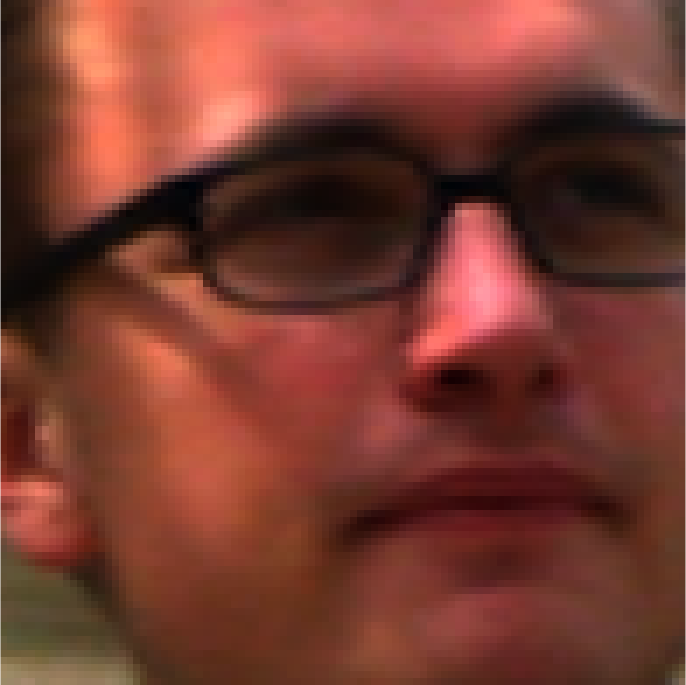}
	\end{tabular}
	\caption{Details from the central view of the light field \texttt{Bikes} (fist and second column from the left),
	\texttt{Chain\_link\_Fence\_2} (third column), \texttt{Flowers} (fourth column), and \texttt{Fountain\_\&\_Vincent} (fifth column) from
	the MMSPG dataset.
	The original low resolution images in the first row are super-resolved by a factor ${\alpha = 2}$ with the methods \cite{mitra_light_2012}
	and GB-SQ in the second and third rows, respectively.}
	\label{fig:mmspg2}
\end{figure*}

\subsection{Light field camera experiments}

We test our algorithm also on the MMSPG dataset \cite{rerabek_new_2016}, where real world scenes are captured with a hand held
\textit{Lytro ILLUM} camera \cite{lytro_inc}.
The super-resolution task happens to be very challenging, as the views in each light field are characterized by a very low resolution and
contain artifacts due to both the uncontrolled light conditions and the demosaicking process.
Moreover, the Lambertian assumption is not always met.
In the tests we keep the parameter setup described in Section~\ref{subsec:exp_settings}, included the ${[-6,6]}$ disparity range.
However, no PSNR is available as the light fields are directly super-resolved.
In Figure~\ref{fig:mmspg2} we provide five examples of light field views super-resolved by a factor ${\alpha = 2}$ with GB-SQ and the method
in \cite{mitra_light_2012}, as the latter represents GB's main competitor in the tests of Section~\ref{subsec:reconstruction_results}.
Consistently with the previous experiments, at depth discontinuities the method in \cite{mitra_light_2012} leads to unpleasant artifacts,
while GB-SQ preserves the sharp transitions.

To conclude, our experiments over three datasets show that the proposed super-resolution algorithm GB has some remarkable
reconstruction properties that make it preferable over its considered competitors.
First, its reconstructed light fields exhibit a better visual quality, often confirmed numerically by the PSNR measure.
In particular, GB leads to sharp edges while avoiding the unpleasant artifacts due to depth discontinuities.
Second, it provides an homogeneous and consistent reconstruction of all the views in the light field, which is a fundamental requirement
for light field applications.
Third, it is more robust than the other considered methods in those scenarios where some objects in the scene exceed the assumed
disparity range, as it may be the case in practice (e.g., in the MMSP dataset), where there is no control on the scene.

\section{Conclusions} \label{sec:conclusions}

We developed a new light field super-resolution algorithm that exploits the complementary information encoded in the different views
to augment their spatial resolution, and that relies on a graph to regularize the target light field.
We showed how to construct the warping matrices necessary to broadcast the complementary information in each view to the whole light field.
In particular, we showed that coupling an approximate warping matrix construction strategy with a graph regularizer that enforces the light field
structure can avoid to carry out an explicit, and costly, disparity estimation step on each view.
We also showed how to extract the warping matrices directly from the graph when computation needs to be kept at the minimum.
Finally, we showed that the proposed algorithm reduces to a simple quadratic problem, that can be solved efficiently with standard
convex optimization tools.

The proposed algorithm compares favorably to the state-of-the-art light field super-resolution frameworks, both in terms of
PSNR and visual quality.
It provides an homogeneous reconstruction of all the views in the light field, which is a property that is not
present in the other light field super-resolution frameworks \cite{wanner_spatial_2012} \cite{mitra_light_2012}.
Also, although the proposed algorithm is meant mainly for light field camera data, where the disparity range is typically small,
it is flexible enough to handle light fields with larger disparity ranges too.
We also compared our algorithm to a state-of-the-art single-frame super-resolution method based on CNNs \cite{dong_learning_2014},
and showed that taking the light field structure into account allows our algorithm to recover finer details and most importantly avoids
the reconstruction of a set of geometrically inconsistent high resolution views.

\section*{Acknowledgements}
We express our thanks to the Swiss National Science Foundation, which supported this work within the project NURIS, and also to
Prof. Christine Guillemot (INRIA Rennes) for the very valuable discussions in the early stages of this work.

\bibliographystyle{IEEEtran}
\bibliography{references}

\end{document}